\documentclass[letterpaper, 10 pt, conference]{ieeeconf} 
\IEEEoverridecommandlockouts
\usepackage{graphicx}
\usepackage{subfig}
\usepackage{amsmath}
\usepackage{mathtools}
\usepackage{bm}
\usepackage{esvect}
\usepackage{gensymb}

\title{\LARGE \bf
Benchmarking Tether-based UAV Motion Primitives 
}

\author{Xuesu Xiao$^{1}$, Jan Dufek$^{1}$, and Robin Murphy$^{1}$
\thanks{$^{1}$Xuesu Xiao, Jan Dufek, and Robin Murphy are with the Department of Computer Science and Engineering,
        Texas A\&M University, College Station, TX 77843
        {\tt\small \{xiaoxuesu, dufek, robin.r.murphy\}@tamu.edu}}%
}

\begin{document}

\maketitle
\thispagestyle{empty}
\pagestyle{empty}

\begin{abstract}

This paper proposes and benchmarks two tether-based motion primitives for tethered UAVs to execute autonomous flight with proprioception only. 
Tethered UAVs have been studied mainly due to power and safety considerations. Tether is either not included in the UAV motion (treated same as free-flying UAV) or only in terms of station-keeping and high-speed steady flight. However, feedback from and control over the tether configuration could be utilized as a set of navigational tools for autonomous flight, especially in GPS-denied environments and without vision-based exteroception. 
In this work, two tether-based motion primitives are proposed, which can enable autonomous flight of a tethered UAV. The proposed motion primitives are implemented on a physical tethered UAV for autonomous path execution with motion capture ground truth. The navigational performance is quantified and compared. 
The proposed motion primitives make tethered UAV a mobile and safe autonomous robot platform. The benchmarking results suggest appropriate usage of the two motion primitives for tethered UAVs with different path plans. 

\end{abstract}

\section{INTRODUCTION}
\label{sec::introduction}
The recent fast development of Unmanned Aerial Vehicles (UAVs) has brought this type of robot into a variety of safety, security, and rescue applications, ranging from marine mass casualty incident response \cite{xiao2017uav, dufek2017visual}, situational awareness enhancement for co-robots team \cite{xiao2017visual}, post-disaster assessment \cite{murphy2016two},  etc. For those applications, one common characteristics is being mission-critical, which means any failure or disruption of the vehicle will result in serious impact on the whole mission. Therefore, safety is emphasized during task execution. One thrust to enable safe and trustworthy UAV application 
is to use a tether. For example, Federal Aviation Administration (FAA) forbids flying UAVs in the United States National Airspace System (NAS) without a Certificate of Authorization (COA), but tethered flight upto 45m is permissible under Federal Aviation Regulation (FAR) \cite{pratt2008use}. Tether could be used as a physical anchor to a fixed point, either a stationary anchor or a moving platform \cite{muttin2011umbilical}, preventing the possibility of UAV ``flying away''. Another important reason for having a tether is to provide extended battery duration \cite{papachristos2014power} and reliable wired communication. Tether could also be used to provide accurate localization of the UAV in GPS-denied environments and where vision-based exteroception is also not available \cite{xiao2018indoor}. 

\begin{figure}
\centering
\subfloat[Indoor Tethered Flight]{\includegraphics[width=0.47\columnwidth]{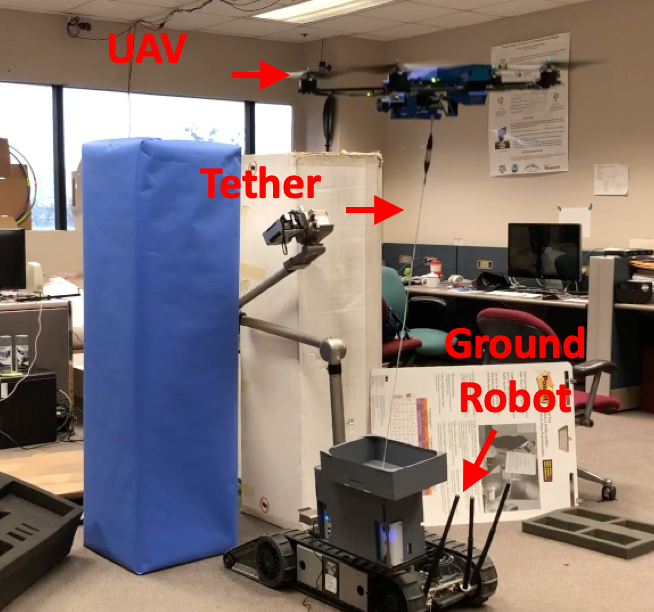}%
\label{fig::indoor}}
\hfill
\subfloat[Outdoor Disaster Environment]{\includegraphics[width=0.5\columnwidth]{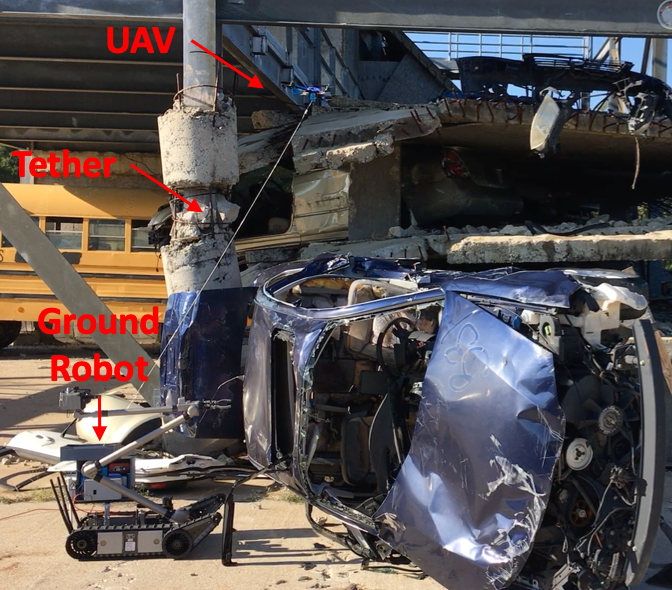}%
\label{fig::outdoor}}
\caption{Co-robots Team with Tethered UAV: ground robot is tele-operated to project human presence in remote environment, but the first-person-view provided by the robot's onboard camera is not sufficient due to the limited field of view and lack of depth perception. A tethered UAV is used as a visual assistant, providing a good external viewpoint to overcome the perceptual limitations of onboard sensors. }
\label{fig::co-robots}
\end{figure}

Fig. \ref{fig::co-robots} shows the co-robots team with tethered UAV developed by our group for robot operations in unstructured or confined environments, using the tethered UAV as a visual assistant at a third-person-view to enhance situational awareness for the tele-operator of an Unmanned Ground Vehicle (UGV) \cite{xiao2018motion, xiao2019autonomous}. Tether is used to match the battery duration of the ground and aerial vehicles in the heterogeneous robot team so as to conduct extended navigation and manipulation tasks in remote environments, such as decommissioning in Fukushima Daiichi nuclear accident. The tether is also a failsafe in case of malfunction, e.g. UAV collision or crash with obstacles. We further derive the relative position of the visual assistant by sensing the tether configuration \cite{xiao2018indoor}. The tethered visual assistant now hovers at a stationary viewpoint or is tele-operated by human operators. Towards developing autonomous behavior of the tethered aerial visual assistant, this research proposes two sets of motion primitives based on the feedback from and control over the tether. These motion primitives are applicable to any tether-based aerial platforms in general and are dependent on proprioception only. This makes tethered UAV platforms more mobile and versatile in mission-critical scenarios, especially for safety, security, and rescue purposes. The performance of the two motion primitives are benchmarked in a motion capture studio, suggesting proper usage scenarios with respect to different path plans. 

The remainder of this paper is organized as follows: Sec. \ref{sec::related_work} discusses related work regarding tethered UAVs. Sec. \ref{sec::approach} proposes the two motion primitives. Sec. \ref{sec::experiments} presents the benchmarking experiments of the two motion primitives with motion capture ground truth. Sec. \ref{sec::discussions} discusses the experimental results and makes suggestions on proper usage of both motion primitives with different path plans. Sec. \ref{sec::conclusions} concludes the paper. 

\section{RELATED WORK}
\label{sec::related_work}
This section reviews the existing literature with regard to tethered UAVs. Despite the advantages thanks to the larger safety margin and extended mission duration, controlling the motion of tethered UAVs has not been extensively investigated. 

Power-over-tether is the main consideration to connect the UAV with a physical tether. Similar to our heterogeneous UAV/UGV co-robots team, \cite{papachristos2014power} proposed a power-tether UAV/UGV team in order to collaboratively navigate partially-mapped environments. Although being motivated by a real-world problem, this work only presented simulation results. Furthermore, the tether was not specifically addressed at all and it was directly assumed that tether could provide extra power, high-bandwidth communication, advanced robustness and stability. The motion of the UAV only considered the tether length constraint and within that constraint the UAV was not treated differently as a free-flying platform. This was the same case for \cite{pratt2008use}, where an unmanned helicopter was used for forensic structural inspection. While the UAV was tele-operated by a pilot, its tether was manually managed by a separate tether manager to control the tether spool and keep enough tension. Other than power and safety considerations, the locomotion of the tethered UAV was not treated differently as a tetherless vehicle. 

In terms of motion, the physical anchor point from the tether can provide extra robustness and stability. \cite{lupashin2013stabilization} utilized the tether in the dynamics model and stabilized the UAV only using a taut tether and inertial sensing. This was the prototype of the tethered UAV platform used in this work. \cite{lupashin2013stabilization} focused on stabilizing the UAV at a stationary position with inertial and tether feedback, but it didn't investigate how UAV motion could be realized through tether-based motion primitives. Even in \cite{schulz2015high}, tether was only used to cancel aerodynamic disturbances in order to achieve high-speed, steady flight in confined environment. The UAV was only able to fly great circles on a sphere centered at the tether anchor point in an open-loop manner. In the above-mentioned work, although tether was considered in the dynamics of the aerial systems, only static or dynamic stabilization of the UAV at or around a stationary point was studied and no attention has been paid to arbitrary locomotion and navigation of a tethered UAV. 

In our previous work, we extended the tether stabilization framework to tether-based localization, with a target application in indoor GPS-denied environment without sufficient onboard computation \cite{xiao2018indoor}. Although the UAV is constantly pulling in order to maintain a taut and straight tether, the increasing gravity with longer tether will inevitably pull the tether down, forming a cantenary instead of a straight line. In this work, we assume that this non-ideal tether deformation is preprocessed and compensated by the localizer proposed in \cite{xiao2018indoor}. 

This work focuses on enabling controlled UAV motion and navigation with tether in 3-D space. Tether serves as the center of our UAV motion primitives: we utilize tether configuration feedback, in the form of tether length, elevation, and azimuth angles, to control the position or velocity components of these variables in order to realize planned 3-D motion in Cartesian space. With the proposed motion primitives, tethered aerial platforms could become more mobile, flexible, and versatile and achieve similar mobility of their tetherless counterpart, while still maintaining the advantages of tethered motion for mission-critical environments. 

\section{TWO MOTION PRIMITIVES}
\label{sec::approach}
This section proposes two tether-based motion primitives, reactive feed-back based position control and model-predictive feedforward velocity control. 

\subsection{Preliminaries}
\begin{figure}[]
\centering
\includegraphics[width=0.9\columnwidth]{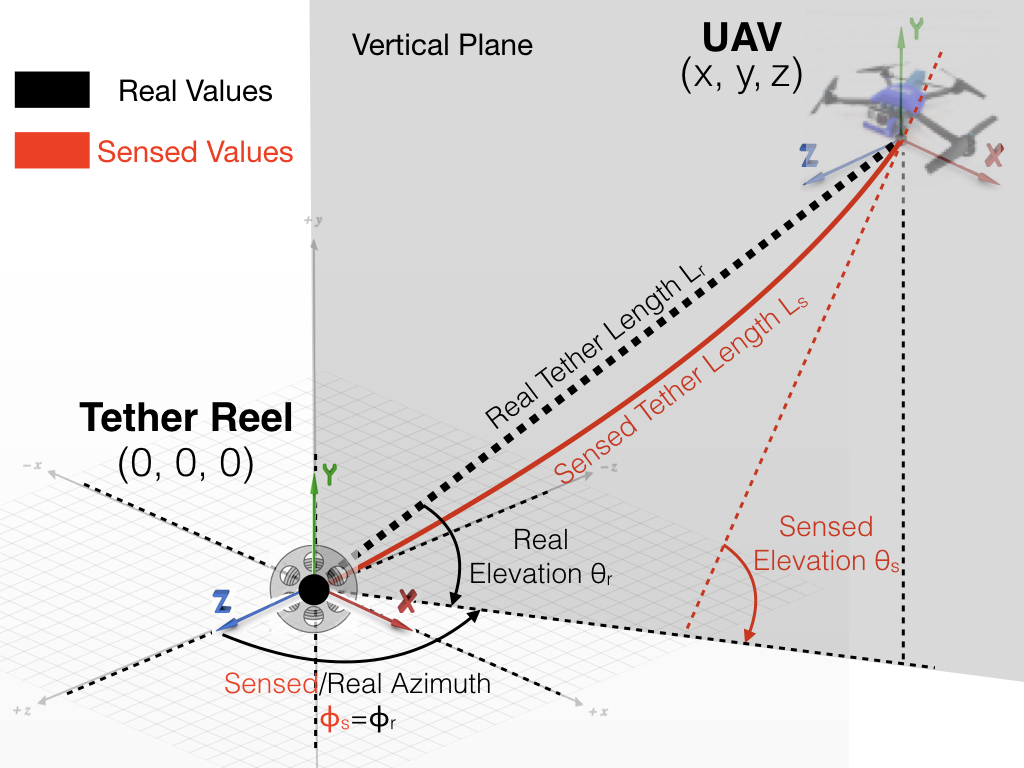}
\caption{The tethered UAV is localized using tether-based sensory feedback including tether length, azimuth, and elevation angles. In this work, we assumes the inaccurate sensed tether feedback is compensated by the localization framework in \cite{xiao2018indoor}. Therefore sensed tether length, azimuth, and elevation angles are regarded as real values. }
\label{fig::real_and_sensed}
\end{figure}

The path plan is given by any type of high-level path planner, in the form of a sequence of 3-dimensional waypoints. The execution of the path is to navigate the tethered UAV along this waypoint sequence in order. \cite{lupashin2013stabilization} presented the controller for elevation and azimuth angles, while tether length is regulated by the tether reel motor. This provides us with our controller input for both motion primitives: change rate of tether length $\dot{L}$, elevation $\dot{\theta}$, and azimuth $\dot{\phi}$. The tethered UAV is by default stabilized around a new equilibrium with regard to a taut tether. The feedback from the tether includes the length computed by tether reel motor encoder, tether elevation and azimuth angle perceived by a piezoelectric deformation sensor mounted on the connecting point between tether and UAV. Tether is expected to be taut and straight, but it is inevitably pulled down by gravity and forms a cantenary, especially when tether is long (shown in red sensed values in Fig. \ref{fig::real_and_sensed}). Using the localizer presented in \cite{xiao2018indoor}, sensed tether-based feedback is corrected and therefore the real values are treated equally as sensed values. This is our sensory feedback of the system: $L_{s}$, $\theta_{s}$, and $\phi_{s}$. 

\subsection{Position Control}
In the position control, we want to utilize the control over change rate of tether length ($\dot{L}$), elevation angle ($\dot{\theta}$), and azimuth angle ($\dot{\phi}$) to realize UAV airframe translational motion in terms of position in 3-D Cartesian space using the onboard position feedback. 

The transformation from the tether-based coordinates to Cartesian coordinates could be expressed as:

\begin{equation}
\label{eqn::polar2euclidean}
\left\{\begin{matrix}
x =& L cos \theta sin \phi \\ 
y =& L sin \theta \\ 
z =& L cos \theta cos \phi
\end{matrix}\right.
\end{equation}

The inverse mapping could be easily derived and gives us the desired tether variables: 

\begin{equation}
\label{eqn::eucledean2polar}
\left\{\begin{matrix}
L_d =& \sqrt{x^2+y^2+z^2}\\ 
\theta_d =& arcsin\frac{y}{\sqrt{x^2+y^2+z^2}}\\ 
\phi_d = &atan2(\frac{x}{z})
\end{matrix}\right.
\end{equation}

Given a 3-D waypoint on a pre-defined path, Eqn. \ref{eqn::eucledean2polar} maps the Cartesian $x$, $y$, and $z$ values into tethered-based $L$, $\theta$, and $\phi$ values. For position control, three independent PD controllers use the three tether input to drive the tether into the desired configuration. Let $e_L$, $e_\theta$, and $e_\phi$ to be the error between desired and sensed value of the three tether variables:

\begin{equation}
\label{eqn::pos_errors}
\vv{\bm{e}}_{\bm{(L, \theta, \phi)}} = \begin{bmatrix} e_L, e_\theta, e_\phi\end{bmatrix}^T= \begin{bmatrix} L_d, \theta_d, \phi_d\end{bmatrix}^T - \begin{bmatrix} L_s, \theta_s, \phi_s\end{bmatrix}^T
\end{equation}

Our control variable $\vv{\bm{u}} = \begin{bmatrix}\dot{L}, \dot{\theta}, \dot{\phi}\end{bmatrix}^T$ are computed by 

\begin{equation}
\label{eqn::pos_controls}
\vv{\bm{u}}
=\vv{\bm{K}}_{\bm{P}}
\vv{\bm{e}}_{\bm{(L, \theta, \phi)}}
+
\vv{\bm{K}}_{\bm{D}}
\dot{\vv{\bm{e}}}_{\bm{(L, \theta, \phi)}}
\end{equation}

where $\vv{\bm{K}}_{\bm{P}}$ and $\vv{\bm{K}}_{\bm{D}}$ are the corresponding proportional and derivative gains: 

\begin{equation}
\label{eqn::kp}
\vv{\bm{K}}_{\bm{P(D)}}=
\begin{bmatrix}
K_{P(D)L}\quad K_{P(D)\theta} \quad K_{P(D)\phi} 
\end{bmatrix}
\end{equation}

Applying $\vv{\bm{u}}$ based on error feedback $\vv{\bm{e}}_{\bm{(L, \theta, \phi)}}$, the system is driven to the desired values. When an acceptance radius is reached around a certain waypoint, the position controller moves on to the next waypoint until the whole sequences is finished. 

\subsection{Velocity Control}
Given the fact that the three PD controllers work independently, it is expected that the position control will achieve unpredictable motion between waypoints. With this in mind, velocity control is proposed to achieve smoother and straighter motion. Based on Eqn. \ref{eqn::polar2euclidean}, the Jacobian matrix of the system could be derived: 

\begin{equation}
\label{eqn::jacobian}
\vv{\bm{\dot{x}}}
=\bm{J} 
\vv{\bm{u}}
\end{equation}

where $\vv{\bm{\dot{x}}}=\begin{bmatrix} \frac{dx}{dt}, \frac{dy}{dt}, \frac{dz}{dt} \end{bmatrix}^T$, $\vv{\bm{u}} = \begin{bmatrix}\dot{L}, \dot{\theta}, \dot{\phi}\end{bmatrix}^T$, and 

\begin{equation}
\label{eqn::j}
\bm{J} = 
\begin{pmatrix}
cos\theta sin\phi & -Lsin\theta sin\phi & Lcos\theta cos\phi\\ 
sin\theta & Lcos\theta & 0\\ 
cos\theta cos\phi & -Lsin\theta cos\phi & -Lcos\theta sin\phi
\end{pmatrix}
\end{equation}

The velocity vector $\vv{\bm{\dot{x}}}$ could be computed by a vector pointing from the current sensed position $\begin{bmatrix} x_s, y_s, z_y \end{bmatrix}^T$ (Eqn. \ref{eqn::polar2euclidean}) to the desired waypoint $\begin{bmatrix} x_d, y_d, z_d \end{bmatrix}^T$: 

\begin{equation}
\label{eqn::norm}
\vv{\bm{\dot{x}}} = 
\alpha \frac{\begin{bmatrix} x_d, y_d, z_d \end{bmatrix}^T - \begin{bmatrix} x_s, y_s, z_y \end{bmatrix}^T}{ \lVert \begin{bmatrix} x_d, y_d, z_d \end{bmatrix}^T - \begin{bmatrix} x_s, y_s, z_y \end{bmatrix}^T \rVert}
\end{equation}

where $\alpha$ is a scalar constant defining the length of the vector, or the absolute speed value of the UAV. So the input $\vv{\bm{u}}$ could be computed by: 

\begin{equation}
\label{eqn::jacobian}
\vv{\bm{u}}
=\bm{J}^{-1} \vv{\bm{\dot{x}}}
\end{equation}

Velocity control aims at the current desired waypoint from the current position at every single time step, and the three control variables in $\vv{\bm{u}}$ are coupled to assure smooth and straight motion. However, when $\theta = 90^\circ$, the Jacobian loses rank and singularity occurs. In fact, even for manual control, the tethered UAV can hardly fly right across the top of the tether reel ($\theta = 90^\circ$). Therefore, $\theta = 90^\circ$ should be avoided when using velocity control. 

\section{EXPERIMENTS}
\label{sec::experiments}
In this section, the two proposed motion primitives are tested in a motion capture (MoCap) studio to quantify their flight performance using two sets of experiments. 

The experiments are conducted in a motion capture studio to capture motion ground truth. In the studio, 6 OptiTrack Flex 13 cameras run at 120 Hz. The 1280$\times$1024 high resolution cameras with a 56\degree~Field of View provide less than 0.5mm positional error and cover a whole 4$\times$4$\times$2.5m space. Eight infrared reflective markers are attached and evenly distributed on all sides of the UAV so that the UAV could be captured even if some of the markers are blocked by the aerial frame itself. 

\begin{figure}[]
\centering
\includegraphics[width=0.8\columnwidth]{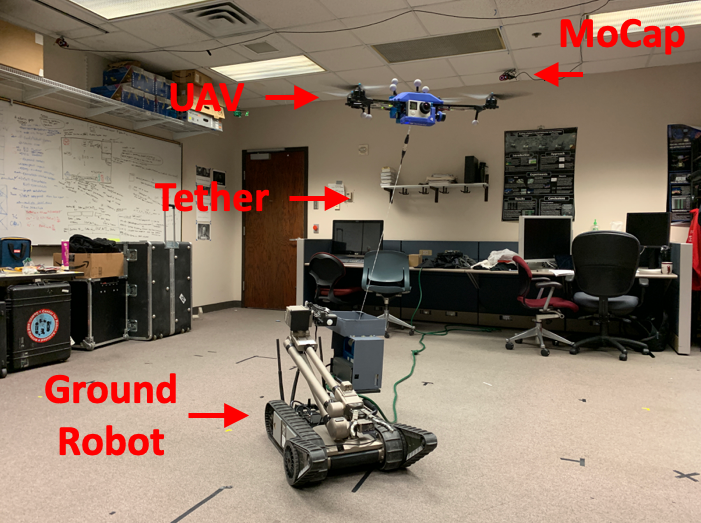}
\caption{Tethered UAV Flying in MoCap Studio}
\label{fig::mocap}
\end{figure}

During the physical tests, the acceptance radius for each waypoint is set to 0.4m. That is, when the UAV is within 0.4m from the current waypoint (localized by onboard sensing only), it is considered that the UAV reaches that particular waypoint and it moves on to the next one. This is the best localization accuracy achievable by the UAV's onboard sensory feedback measured by experiments. Fig. \ref{fig::mocap} shows the tethered UAV flying in the MoCap studio. 

\begin{figure}
\centering
\subfloat[Auxiliary View]{\includegraphics[width=0.5\columnwidth]{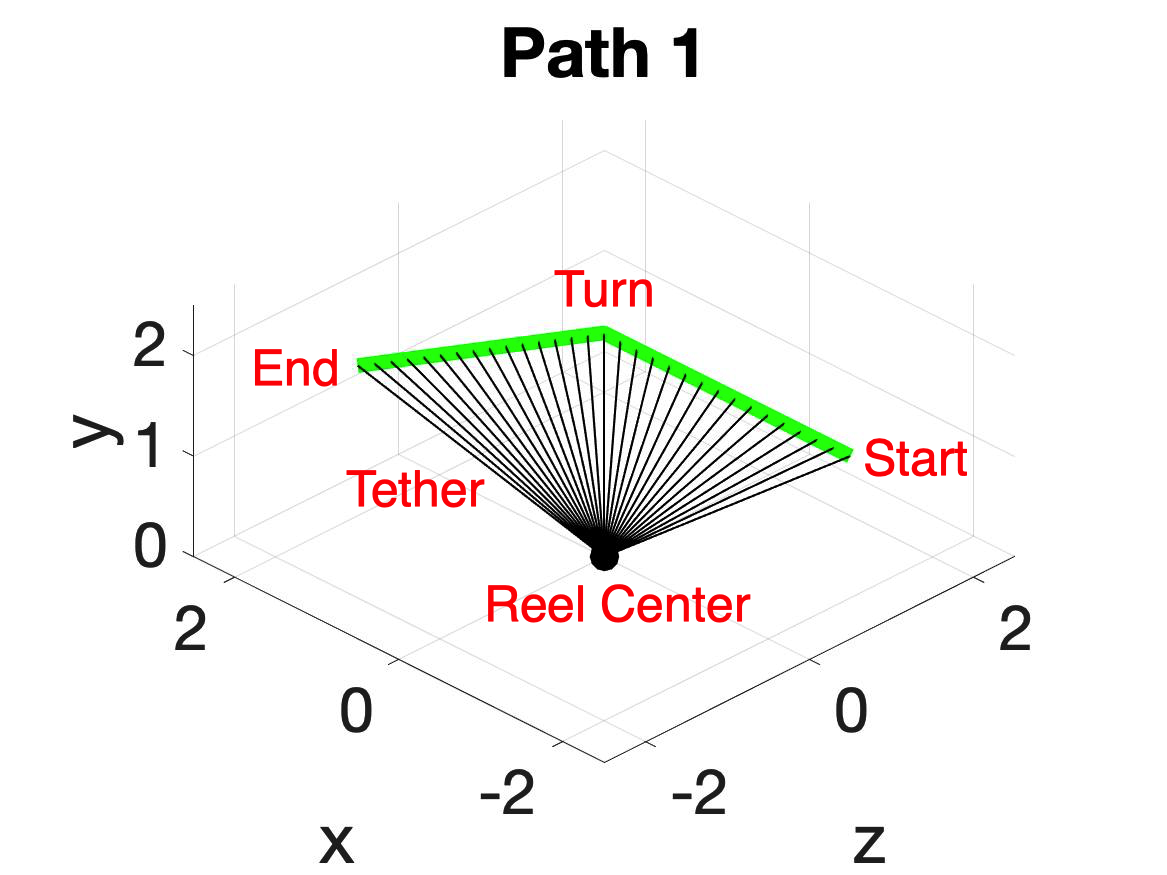}%
\label{fig::auxiliary}}
\subfloat[Front View]{\includegraphics[width=0.5\columnwidth]{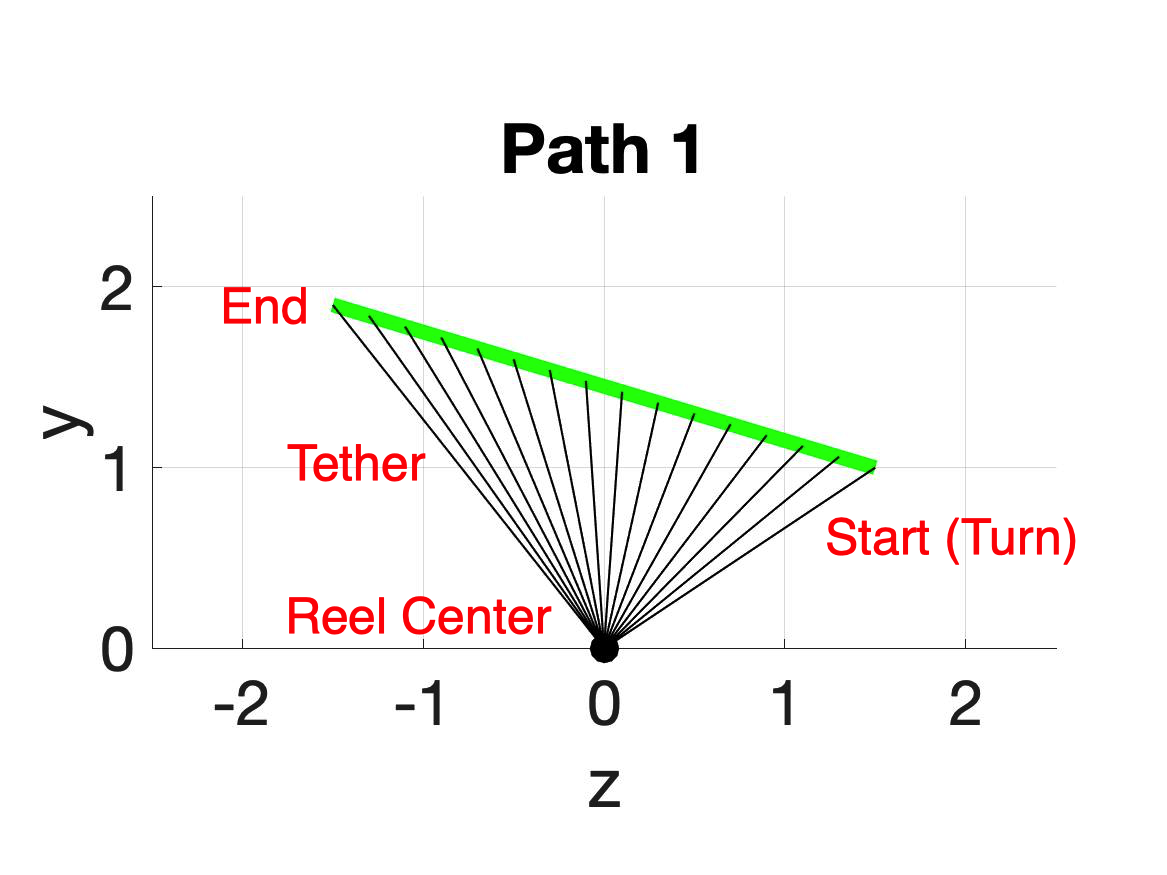}%
\label{fig::front}}\\
\subfloat[Top View]{\includegraphics[width=0.5\columnwidth]{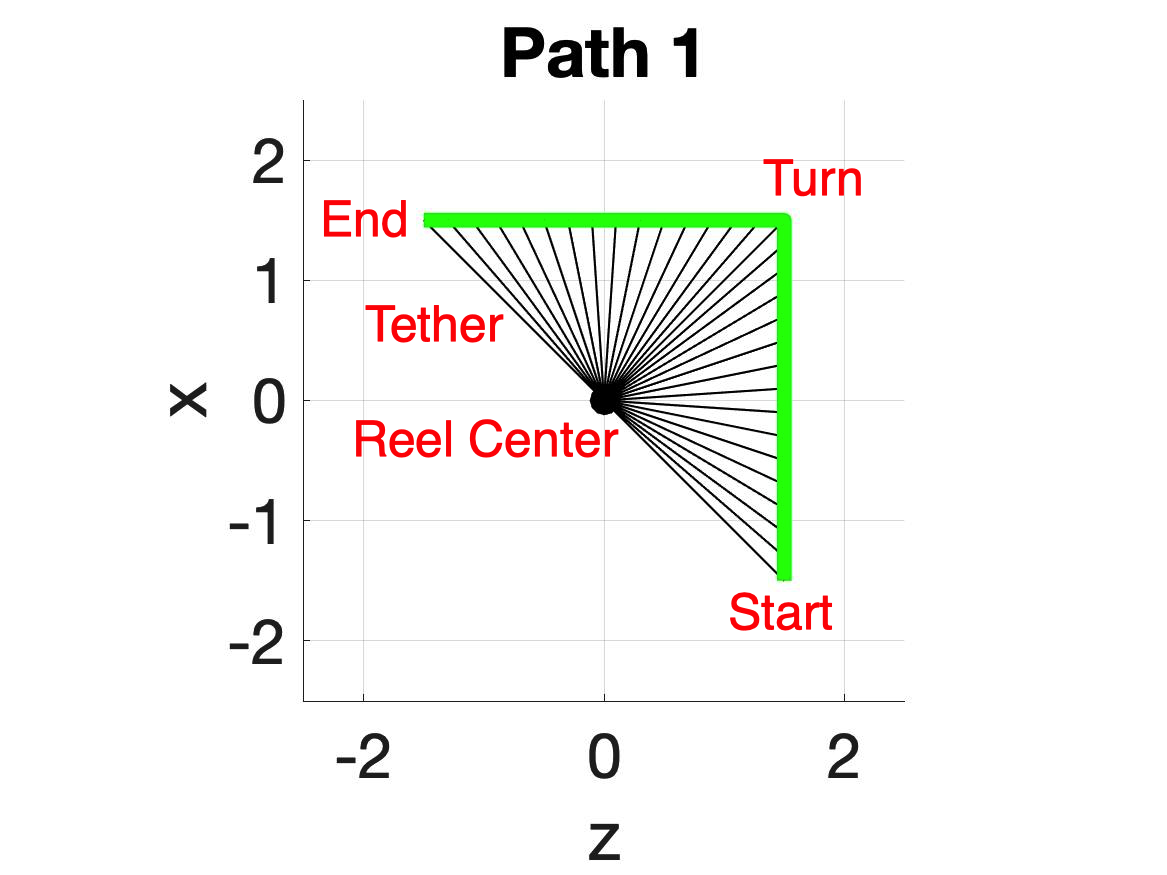}%
\label{fig::top}}
\subfloat[Side View]{\includegraphics[width=0.5\columnwidth]{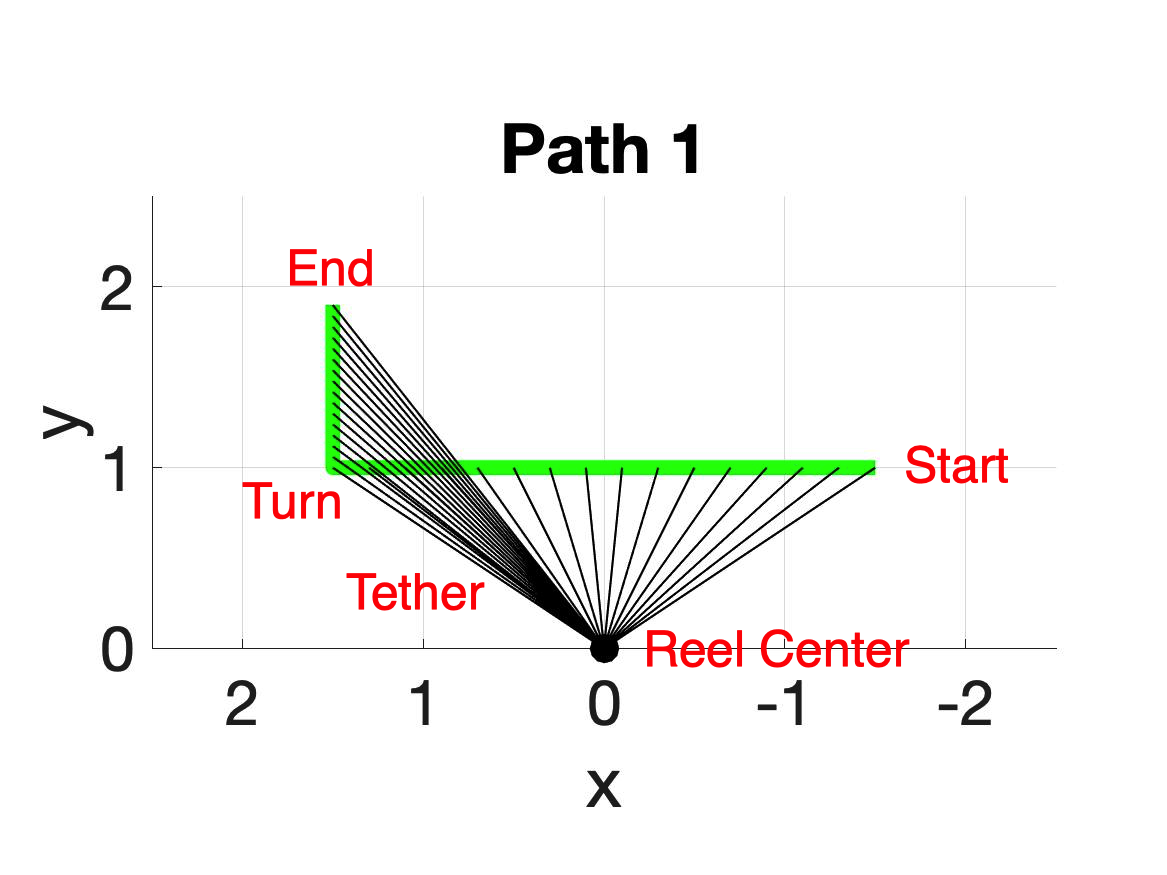}%
\label{fig::side}}
\caption{Different Views for Experiment 1}
\label{fig::path1_views}
\end{figure}

Executing a straight line path may be trivial for free flying UAVs, but the straightness and accuracy of the path execution is of importance to tethered UAVs. In the first set of experiments, we first test a flight path consisting of a 3m horizontal and an ascending straight path (3m projection length on horizontal plane) connected by a 90\degree~turn (Fig. \ref{fig::path1_views}). We test both motion primitives on path plans with five different waypoint densities. That is, from dense to sparse, the intervals between two consecutive waypoints projected in the horizontal plane are 0.2m, 0.5m, 1m, 1.5m, and 3m. Therefore the numbers of waypoints for each path plan are 31, 13, 7, 5, and 3, respectively, denoting the same path. For each waypoint density, six repetitive trials are executed, three of which using position control and other three using velocity control. A second set of experiments is conducted on a straight line path passing above the tether reel center from the first to third quadrant in the x-z plane. This set of experiments shows the inability of velocity control near singularity and the improvement of flight accuracy with denser waypoints using position control. Since the orientation control of the UAV is not the focus of this research, the yaw is not explicitly controlled during path execution. 

\section{Discussions}
\label{sec::discussions}
\begin{figure*}
\centering
\subfloat[Position 0.2m]{\includegraphics[width=0.4\columnwidth]{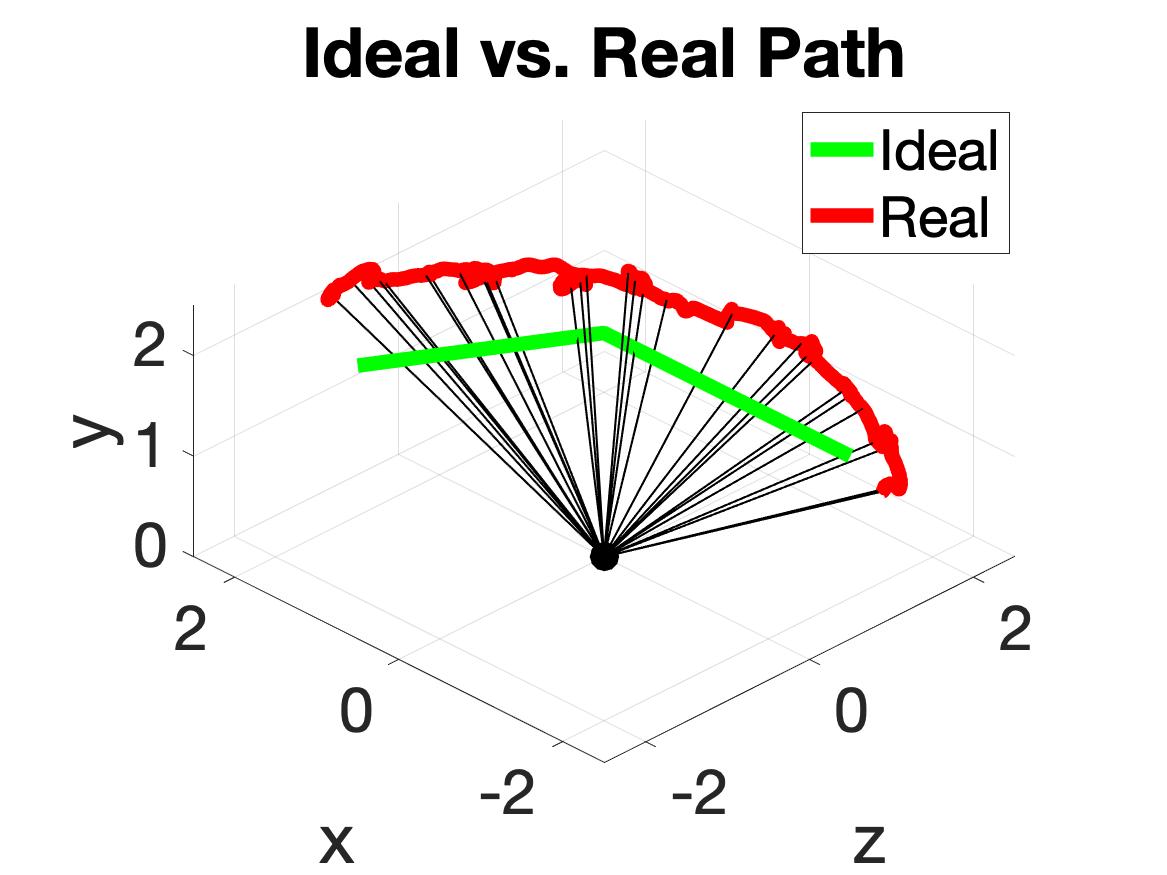}%
\label{fig::pos0.2}}
\subfloat[Position 0.5m]{\includegraphics[width=0.4\columnwidth]{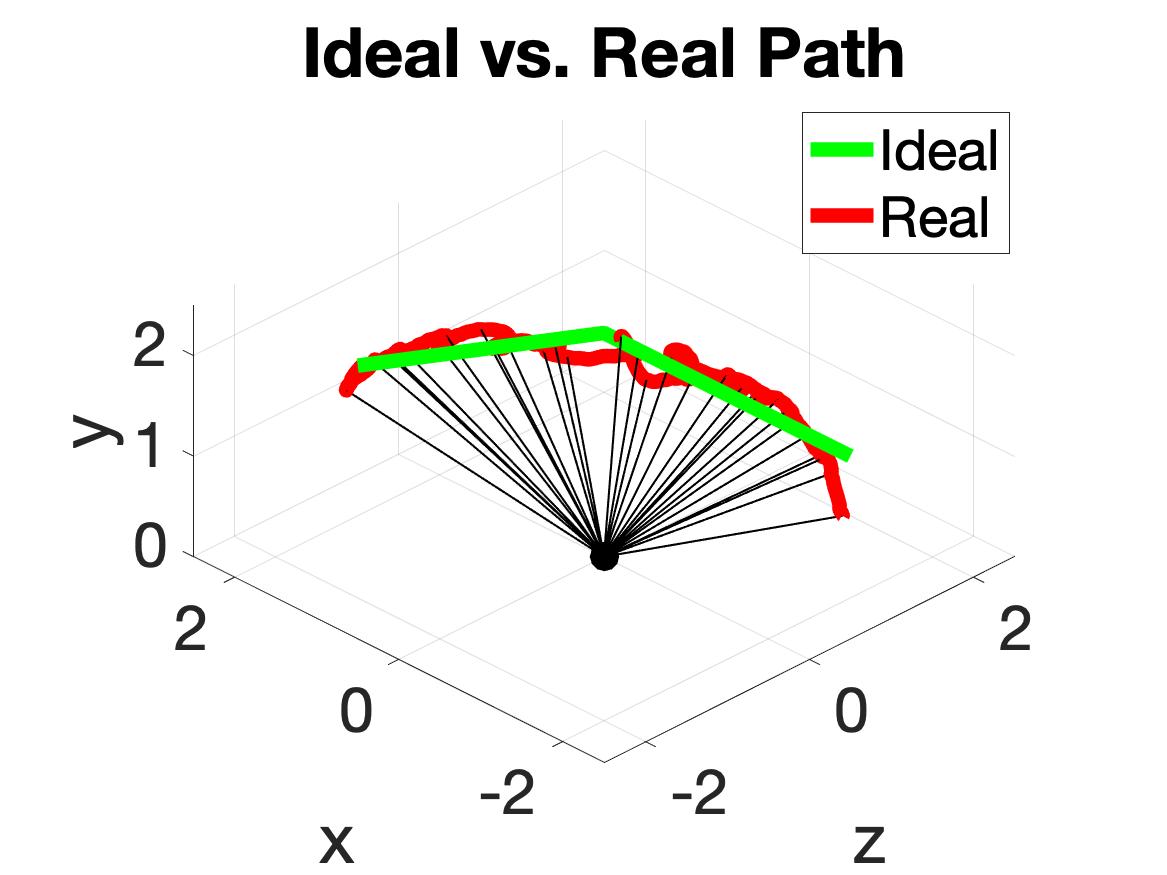}%
\label{fig::pos0.5}}
\subfloat[Position 1m]{\includegraphics[width=0.4\columnwidth]{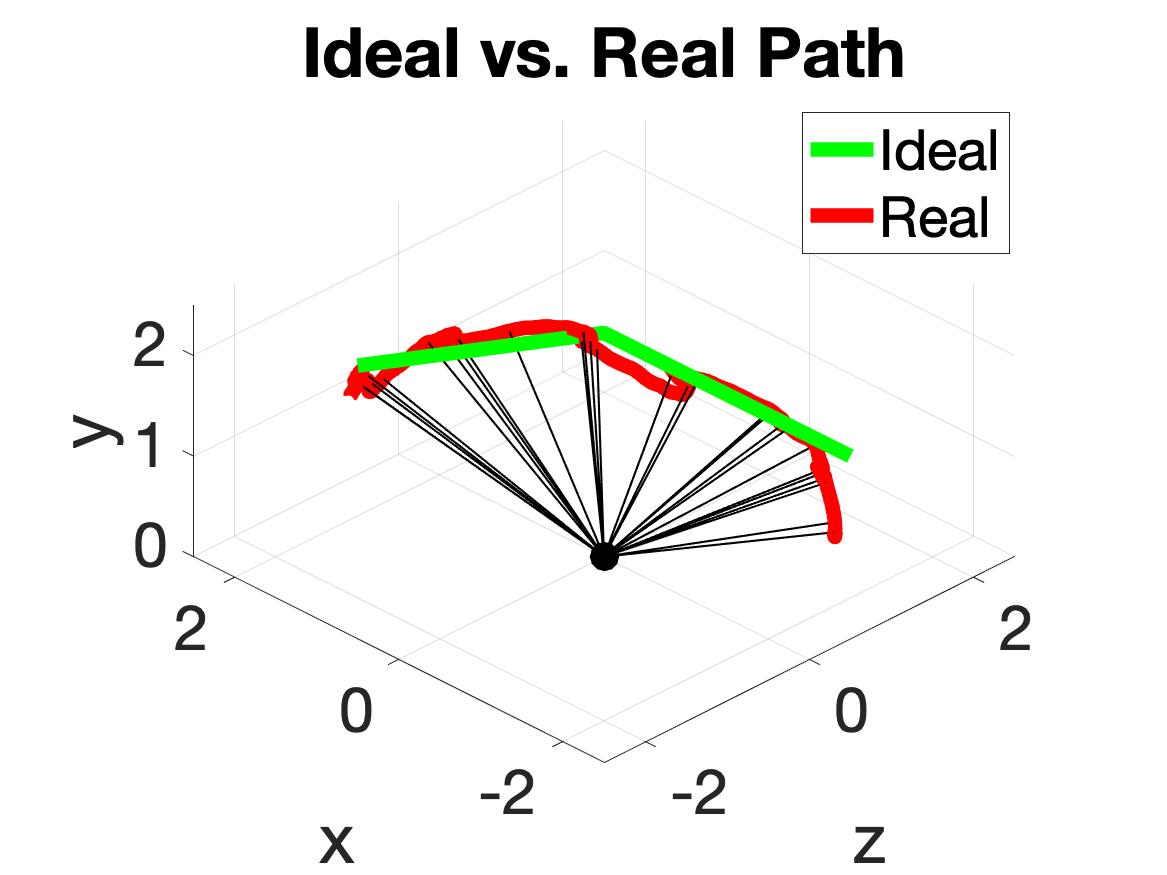}%
\label{fig::pos1}}
\subfloat[Position 1.5m]{\includegraphics[width=0.4\columnwidth]{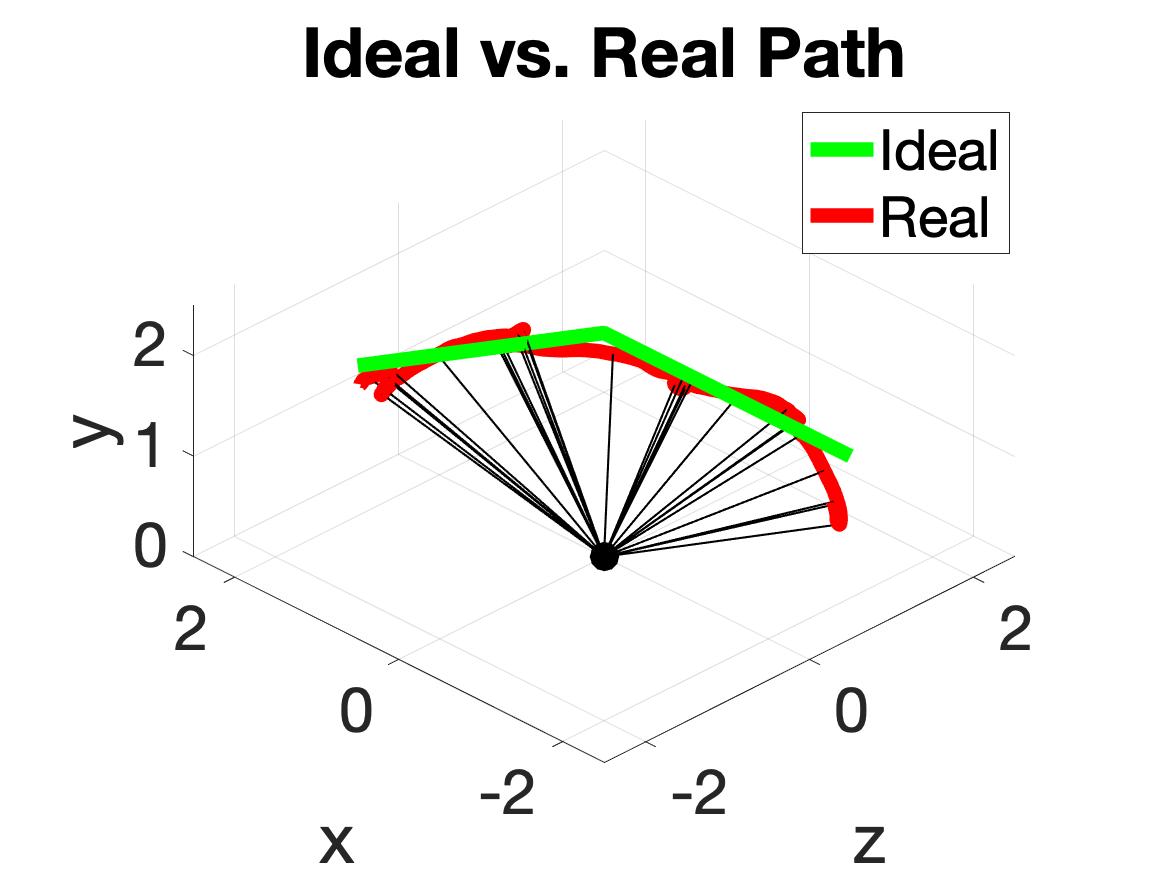}%
\label{fig::pos1.5}}
\subfloat[Position 3m]{\includegraphics[width=0.4\columnwidth]{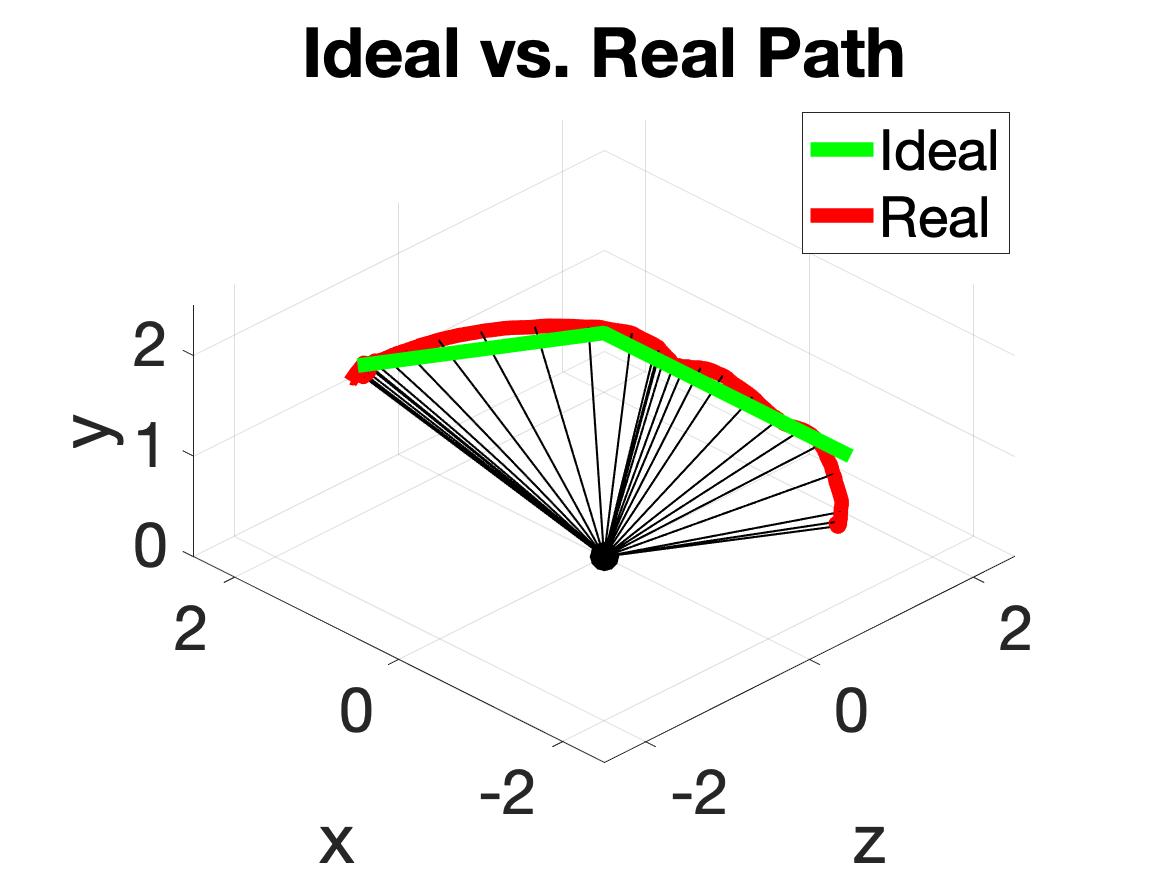}%
\label{fig::pos3}}\\
\subfloat[Velocity 0.2m]{\includegraphics[width=0.4\columnwidth]{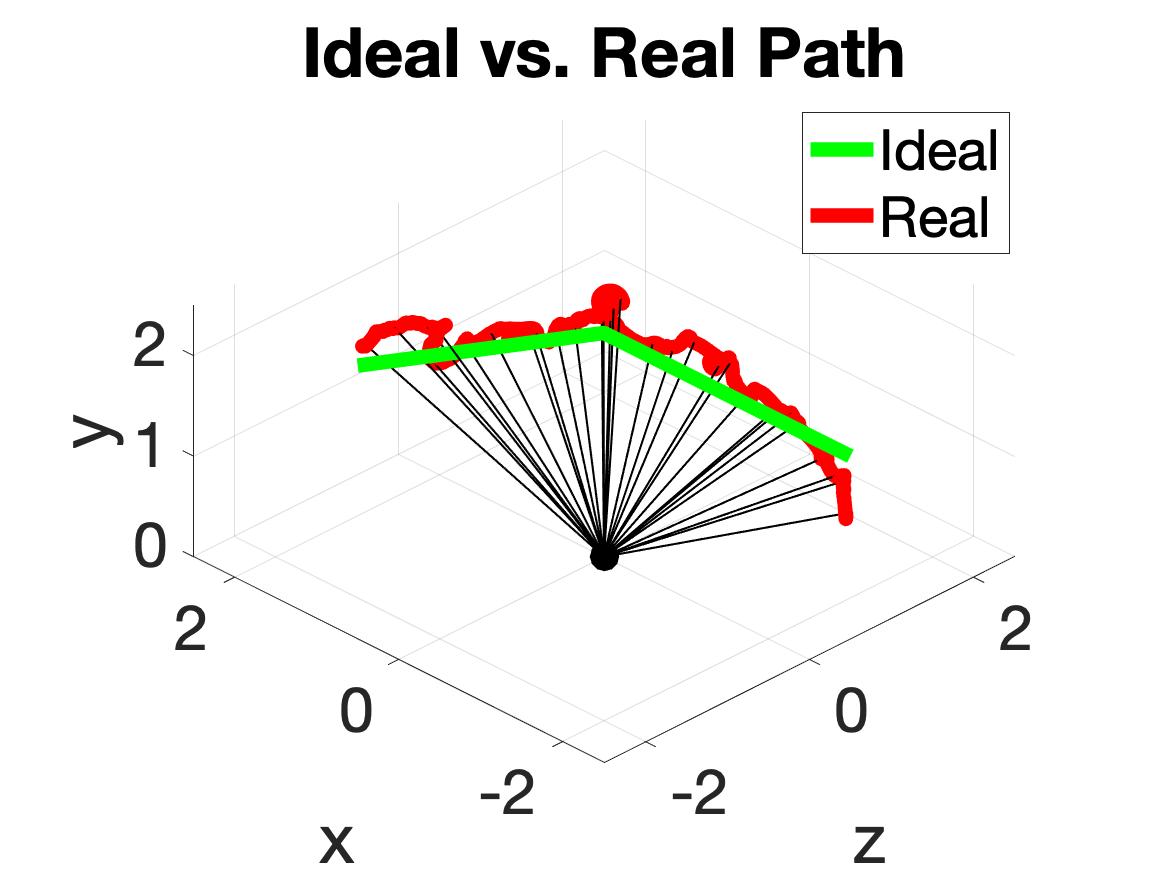}%
\label{fig::vel0.2}}
\subfloat[Velocity 0.5m]{\includegraphics[width=0.4\columnwidth]{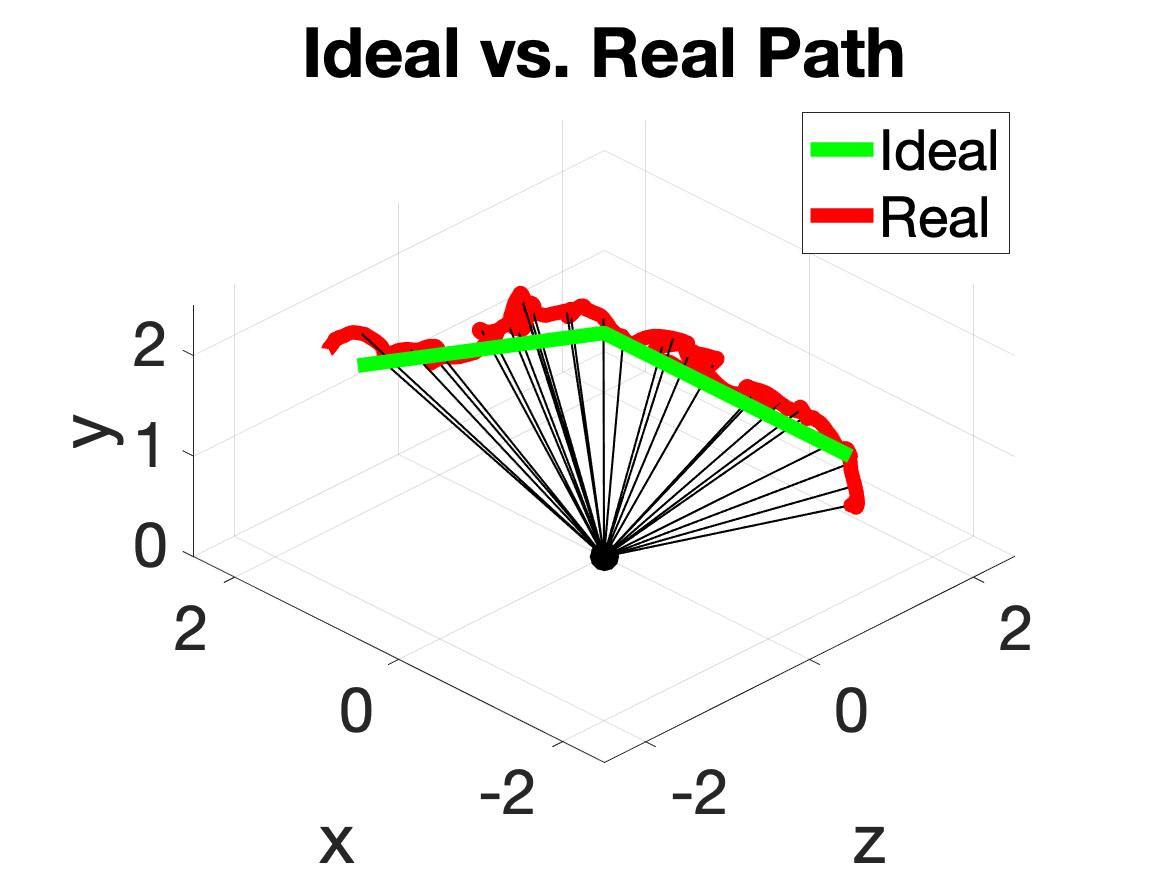}%
\label{fig::vel0.5}}
\subfloat[Velocity 1m]{\includegraphics[width=0.4\columnwidth]{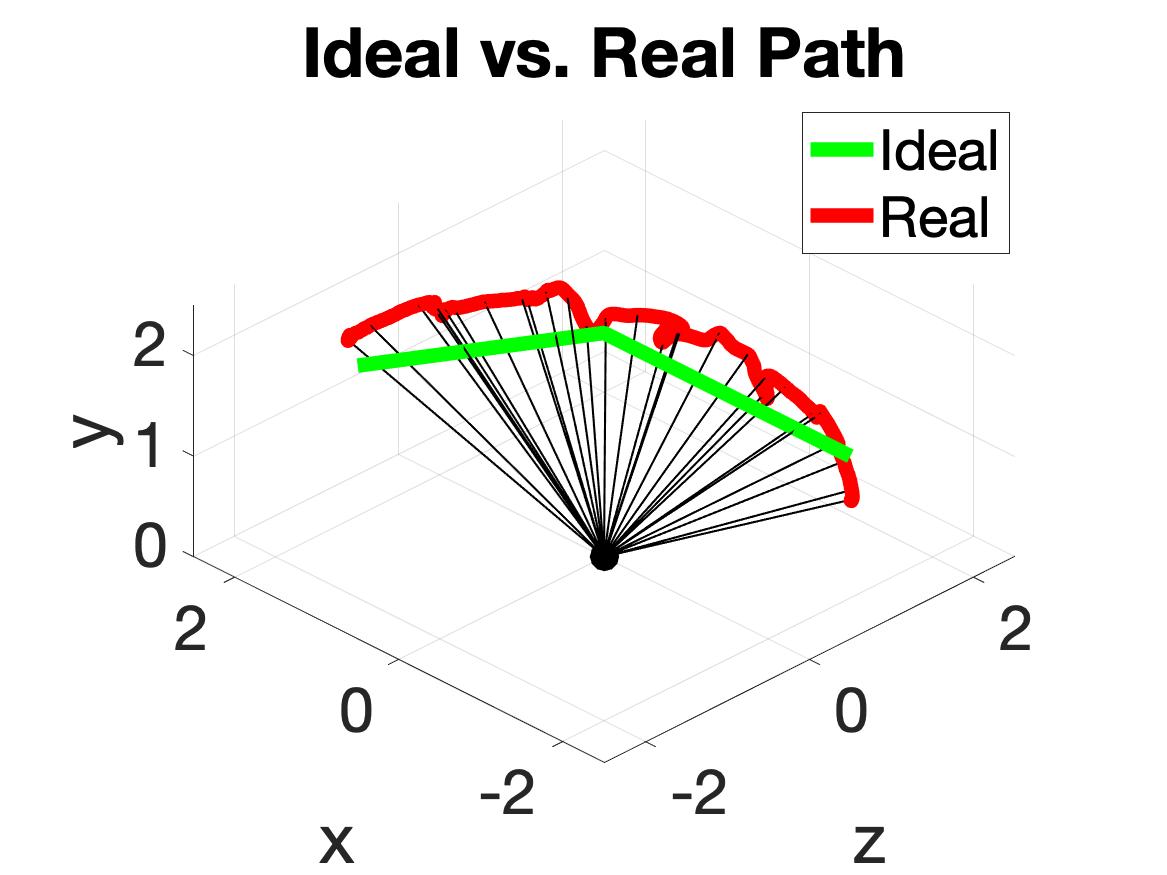}%
\label{fig::vel1}}
\subfloat[Velocity 1.5m]{\includegraphics[width=0.4\columnwidth]{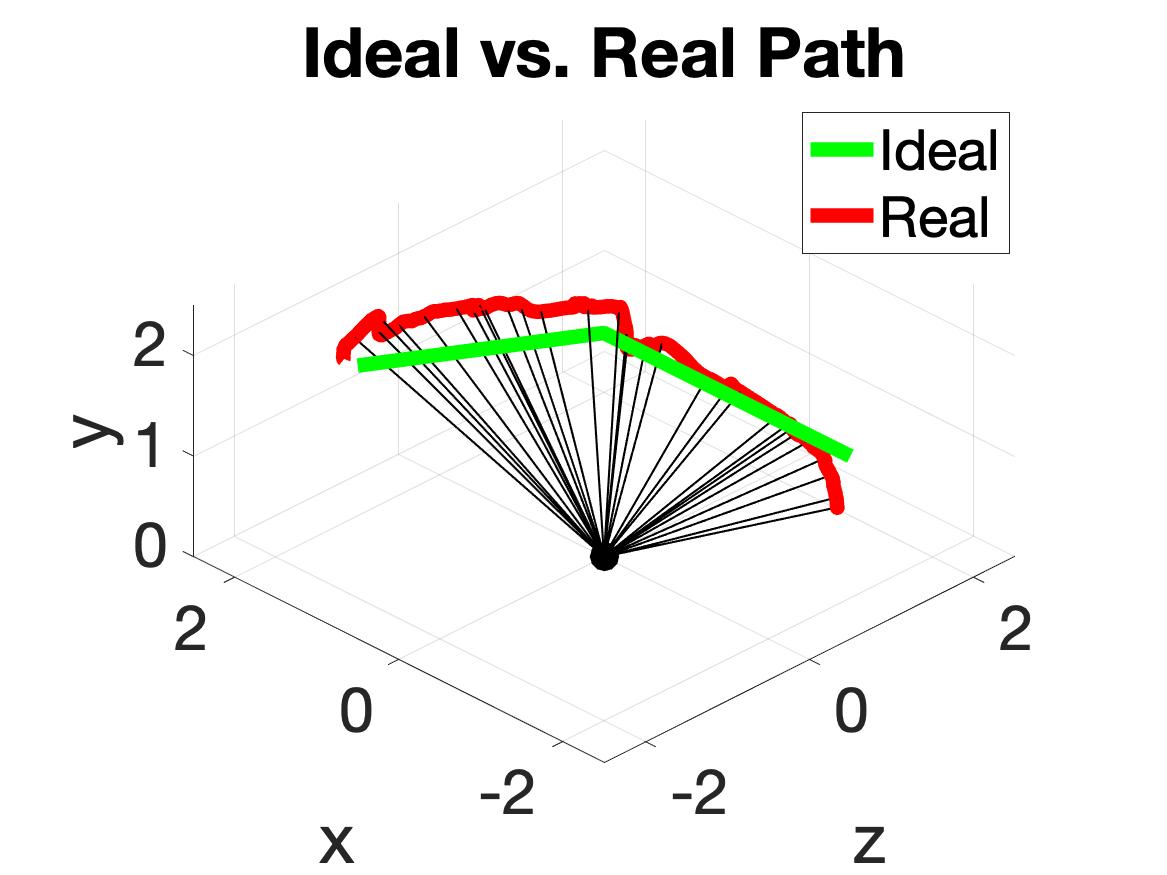}%
\label{fig::vel1.5}}
\subfloat[Velocity 3m]{\includegraphics[width=0.4\columnwidth]{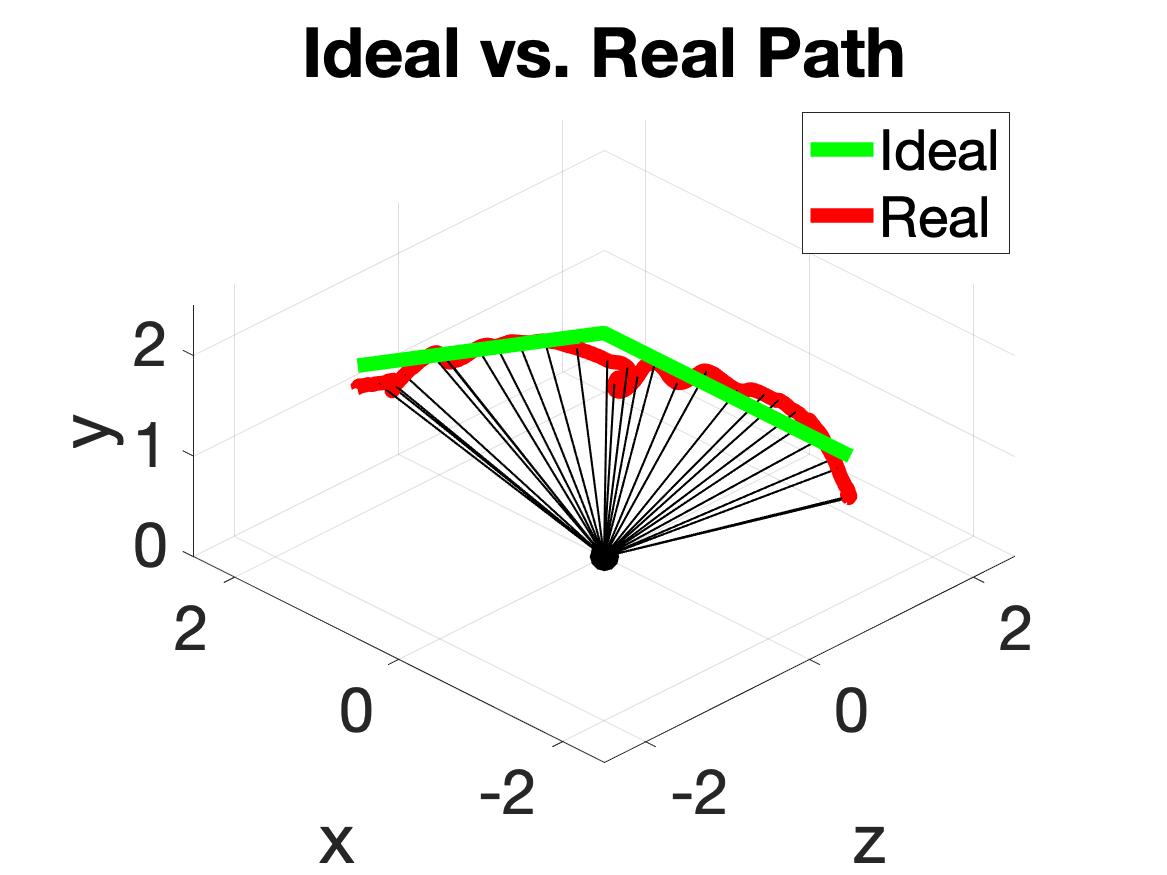}%
\label{fig::vel3}}
\caption{Experimental Results}
\label{fig::experiments}
\end{figure*}

Among the total 30 trials, one example trial is randomly selected for each density and each motion primitive and is shown in Fig. \ref{fig::experiments}. 

\begin{figure}[]
\centering
\includegraphics[width=0.7\columnwidth]{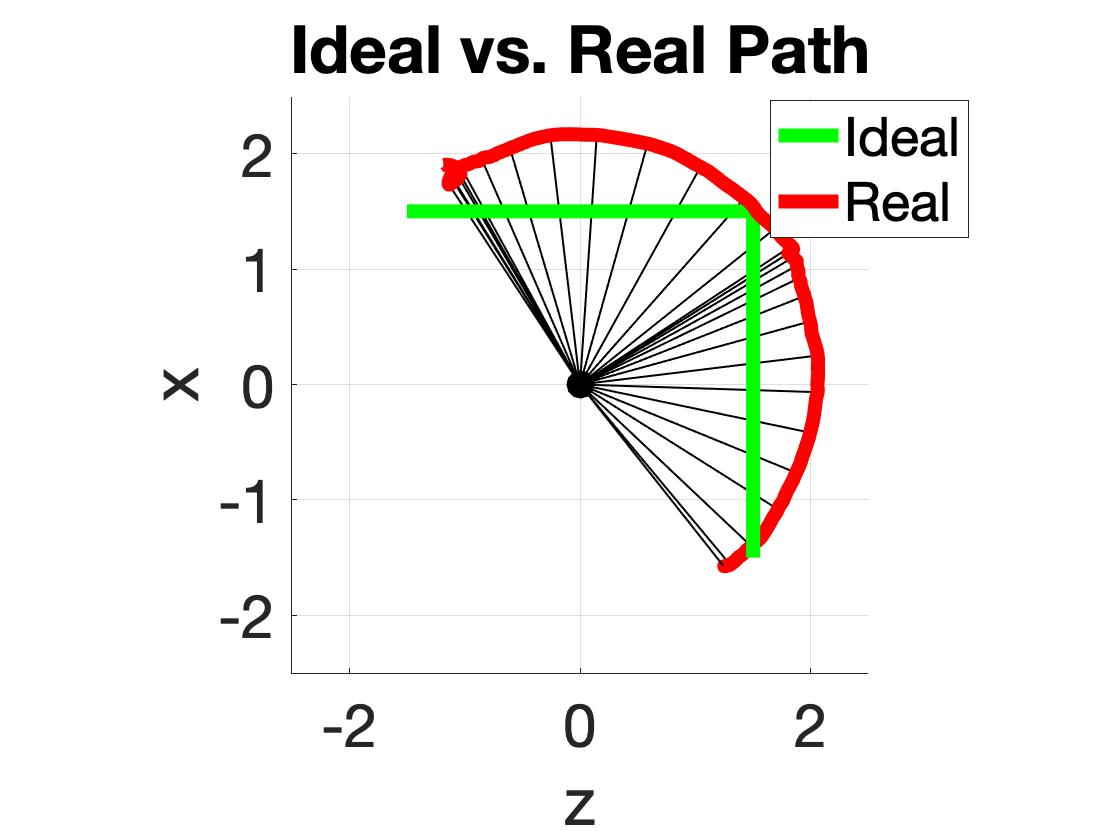}
\caption{Top View of Position Control on 3m Interval}
\label{fig::pos_3_1_up}
\end{figure}

\begin{figure}[]
\centering
\includegraphics[width=0.7\columnwidth]{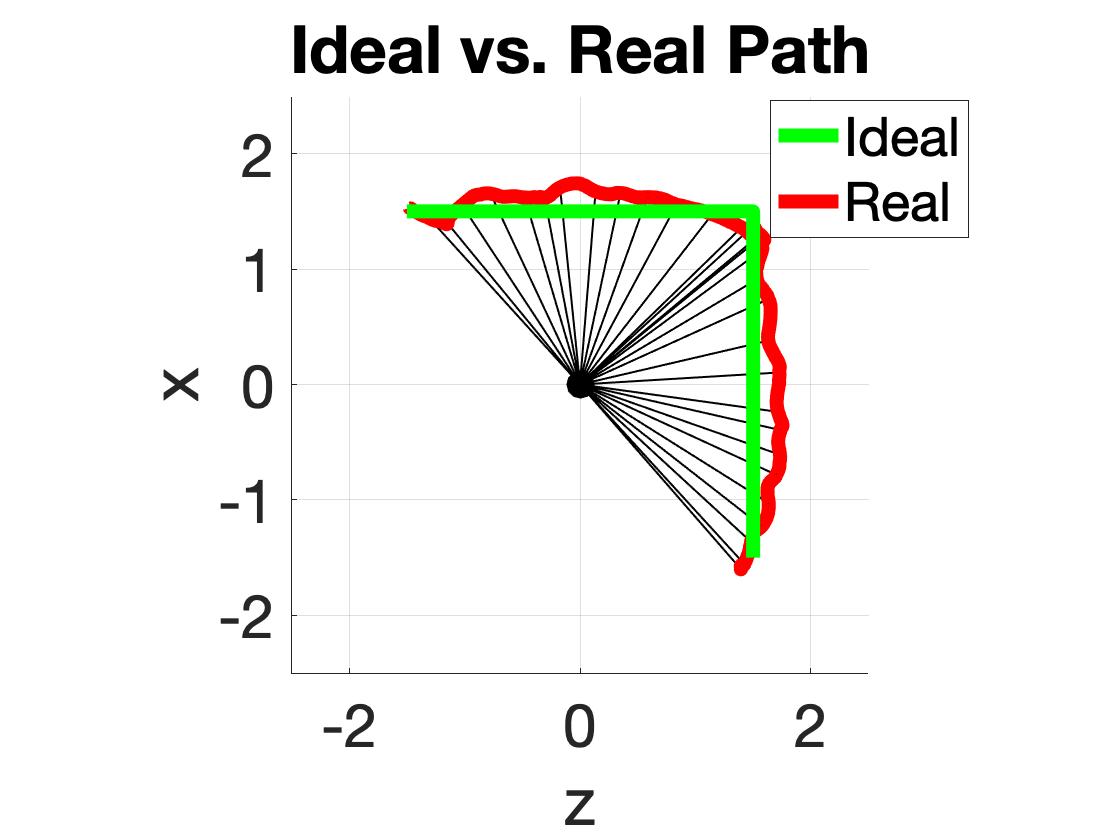}
\caption{Top View of Velocity Control on 3m Interval}
\label{fig::vel_3_3_up}
\end{figure}

One closer look into the two motion primitives on 3m interval is shown in Fig. \ref{fig::pos_3_1_up} and Fig. \ref{fig::vel_3_3_up}. These clearly demonstrate the problem with position control on path plans with sparse waypoints: since the three tether variables are controlled independently, the trajectory between two consecutive waypoints are non-deterministic. In Fig. \ref{fig::pos_3_1_up} and Fig. \ref{fig::vel_3_3_up}, only three waypoints are used to defined the start (lower right), turn (upper right), and end (upper left) point of the path. Apparently the first two points have the same tether length, therefore the position controller does not change the tether length at all and makes an arc-like trajectory instead of a straight line (Fig. \ref{fig::pos_3_1_up}). The end point has slightly longer tether length due to the increase in elevation, and the UAV executes a similar path. This does not happen in velocity control (Fig. \ref{fig::vel_3_3_up}) due to the coordination through system Jacobian. It is expected that position control may perform better with dense waypoints. 

\begin{figure}[]
\centering
\includegraphics[width=0.7\columnwidth]{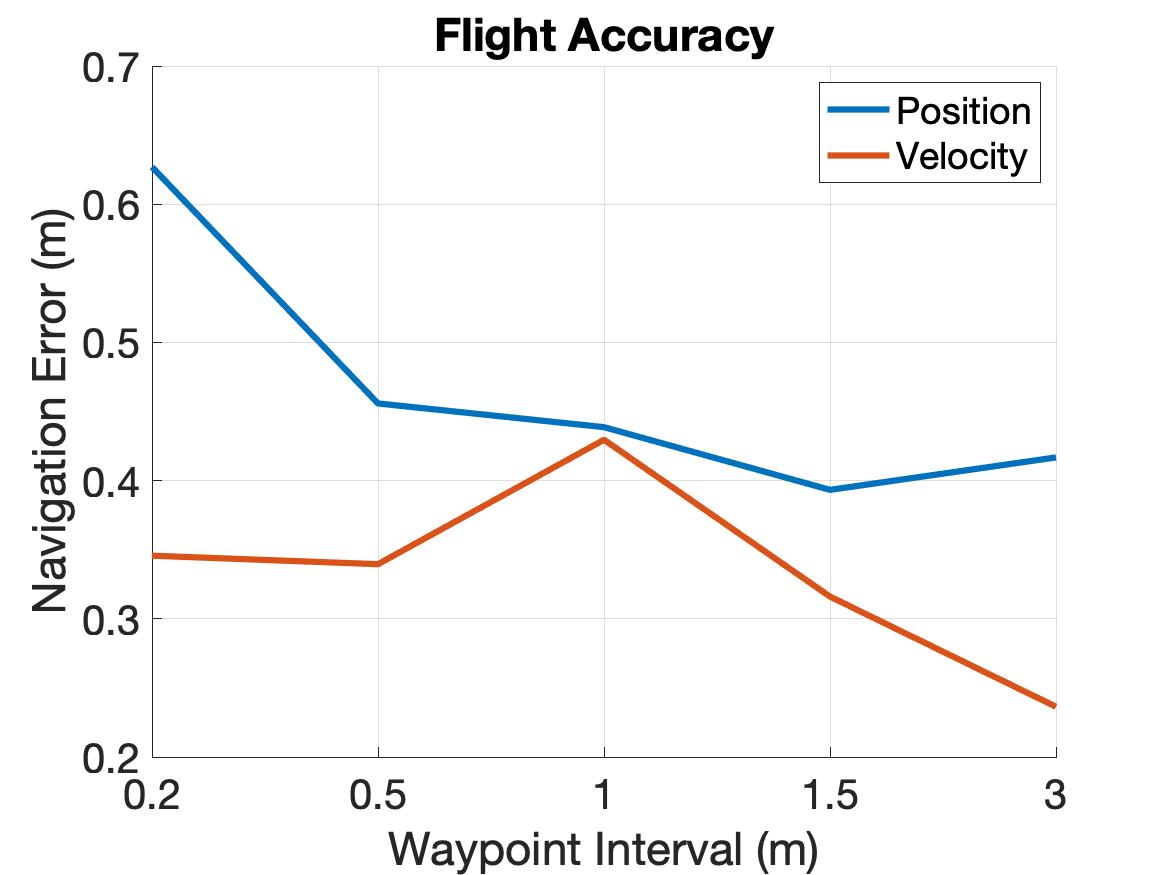}
\caption{Flight Accuracy in Terms of Cross Track Error}
\label{fig::flight_accuracy}
\end{figure}

\begin{figure}[]
\centering
\includegraphics[width=0.7\columnwidth]{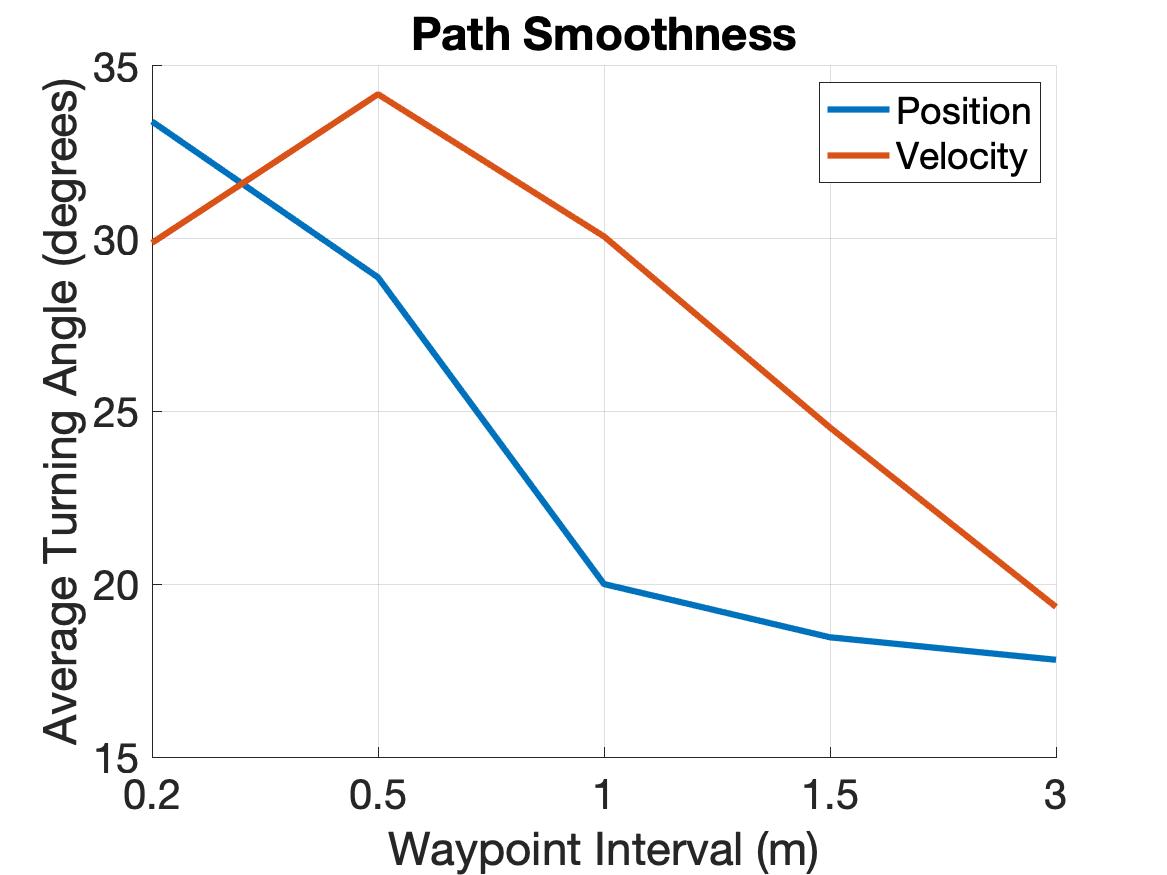}
\caption{Path Smoothness in Terms of Angular Difference}
\label{fig::path_smoothness}
\end{figure}

The average flight accuracy (error) and path smoothness are analyzed in Fig. \ref{fig::flight_accuracy} and Fig. \ref{fig::path_smoothness} . Flight accuracy is defined as the cross track error between the real and ideal trajectory. Path smoothness is the average angular difference between two vectors connecting two pairs of consecutive waypoints. Path is smoother with sparser waypoints for both motion primitives. This is because when executing sparse waypoints, both motion primitives are aiming at a farther waypoint, instead of focusing on some waypoint in the vicinity. The ``short-sightedness'' caused by dense waypoints will introduce instability to the controller, such as overshoot by trying too hard to converge to the ideal path. With sparse waypoints, on the other hand, both controllers act using ``line-of-sight'', aiming at the path ahead of the UAV and avoiding over-compensation. Flight accuracy for velocity control in Fig. \ref{fig::flight_accuracy} is not very sensitive to waypoint density. One surprising result of flight accuracy is for position control: instead of increasing error with sparser waypoints, error actually decreases. Upon examination of the captured trajectories, it is found out that the expected error caused by the independent control of the three tether variables (Fig. \ref{fig::pos_3_1_up}) is in the same range as the UAV flight tolerance 0.4m. Therefore, even with a path plan with dense waypoints, the expected better accuracy is actually canceled by the large tolerance value around all waypoints. 

\begin{figure*}
\centering
\subfloat[Position 0.2m]{\includegraphics[width=0.4\columnwidth]{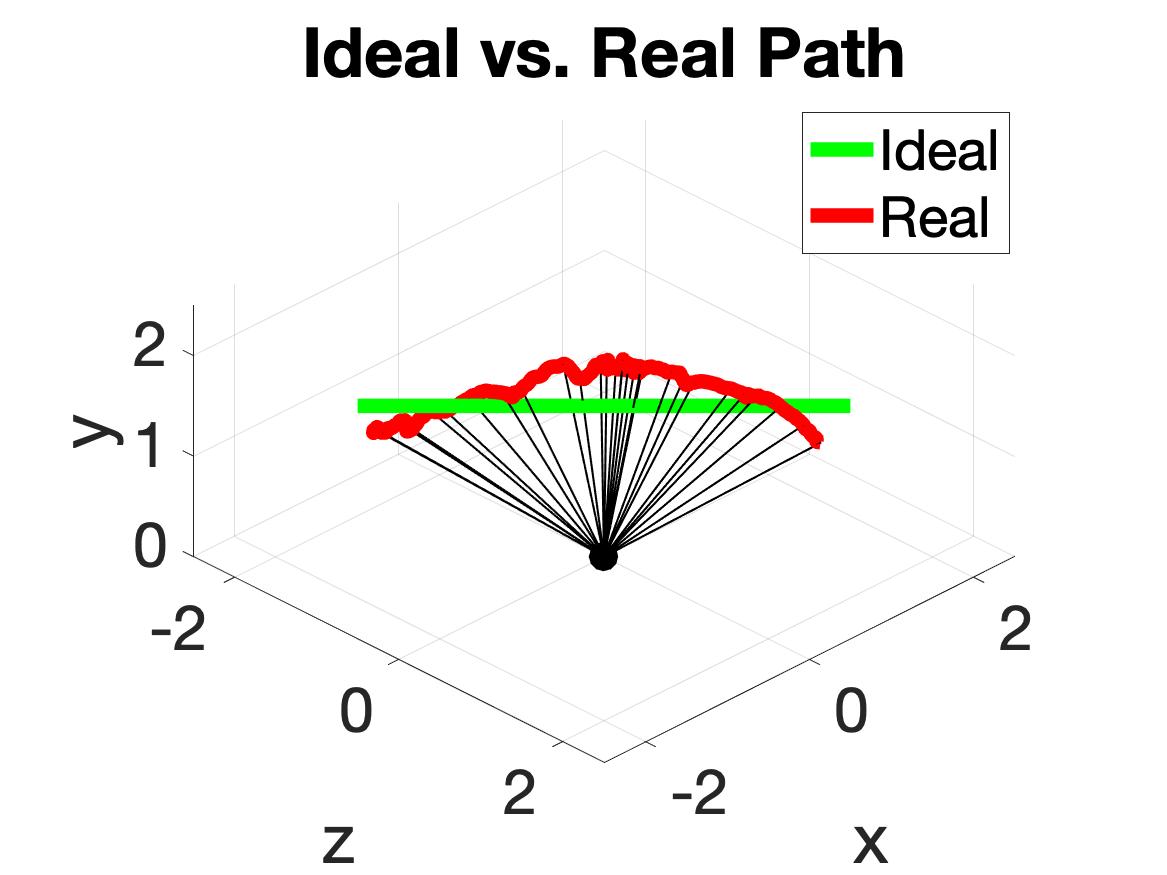}%
\label{fig::pos_02_1_2}}
\subfloat[Position 0.5m]{\includegraphics[width=0.4\columnwidth]{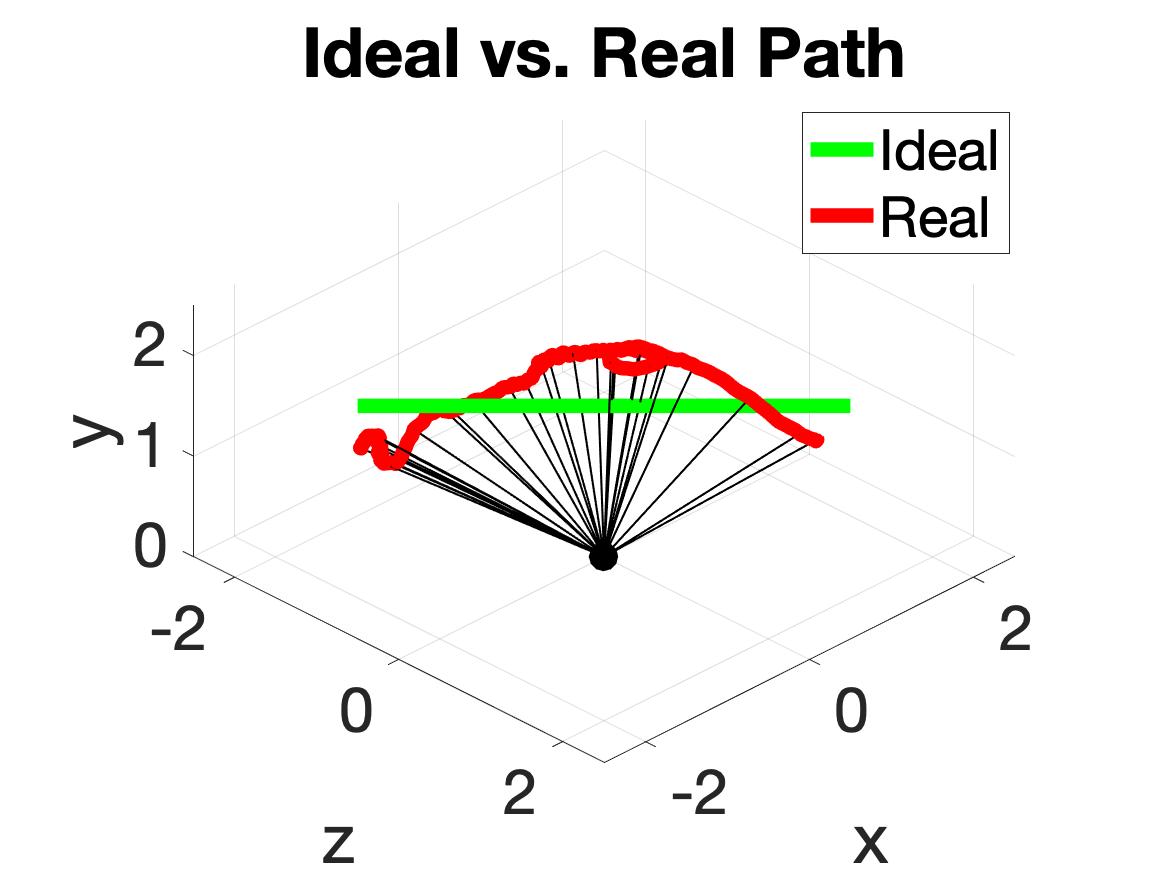}%
\label{fig::pos_05_2_2}}
\subfloat[Position 1m]{\includegraphics[width=0.4\columnwidth]{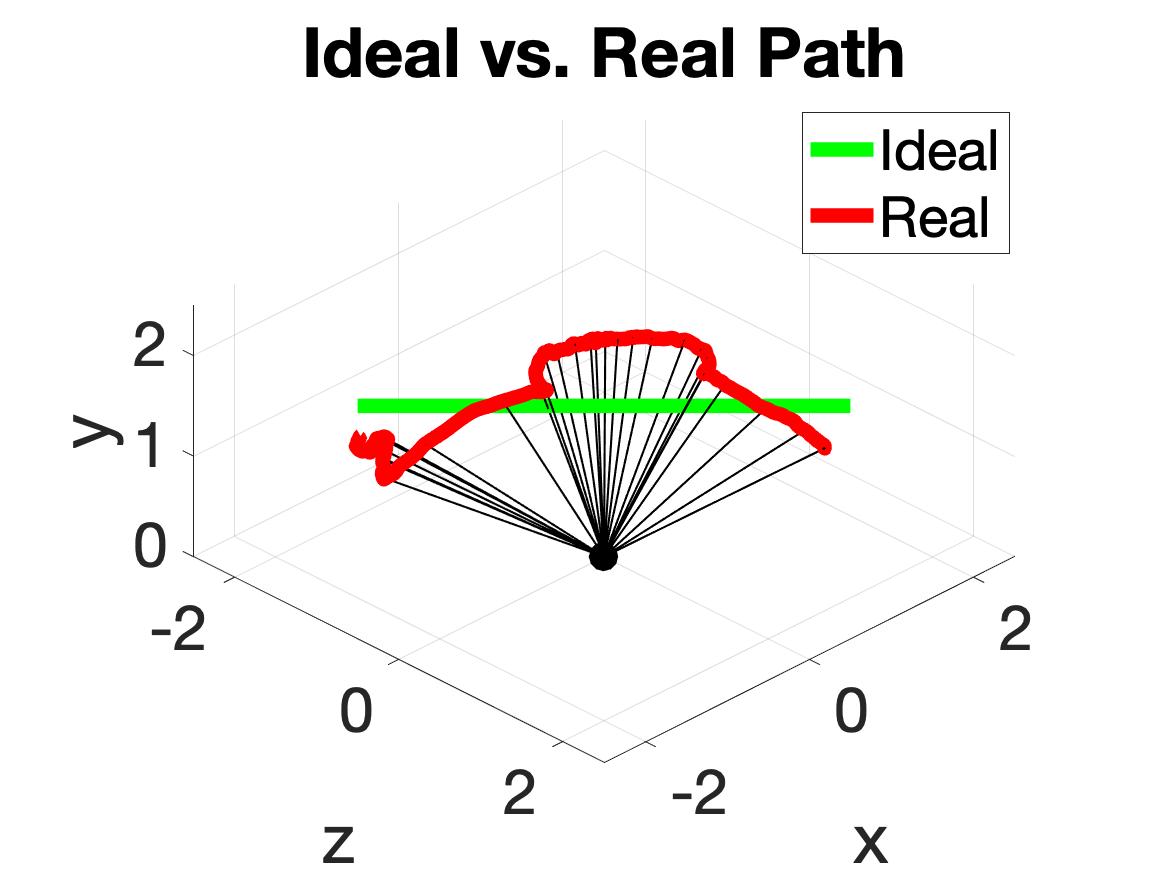}%
\label{fig::pos_1_1_2}}
\subfloat[Position 1.5m]{\includegraphics[width=0.4\columnwidth]{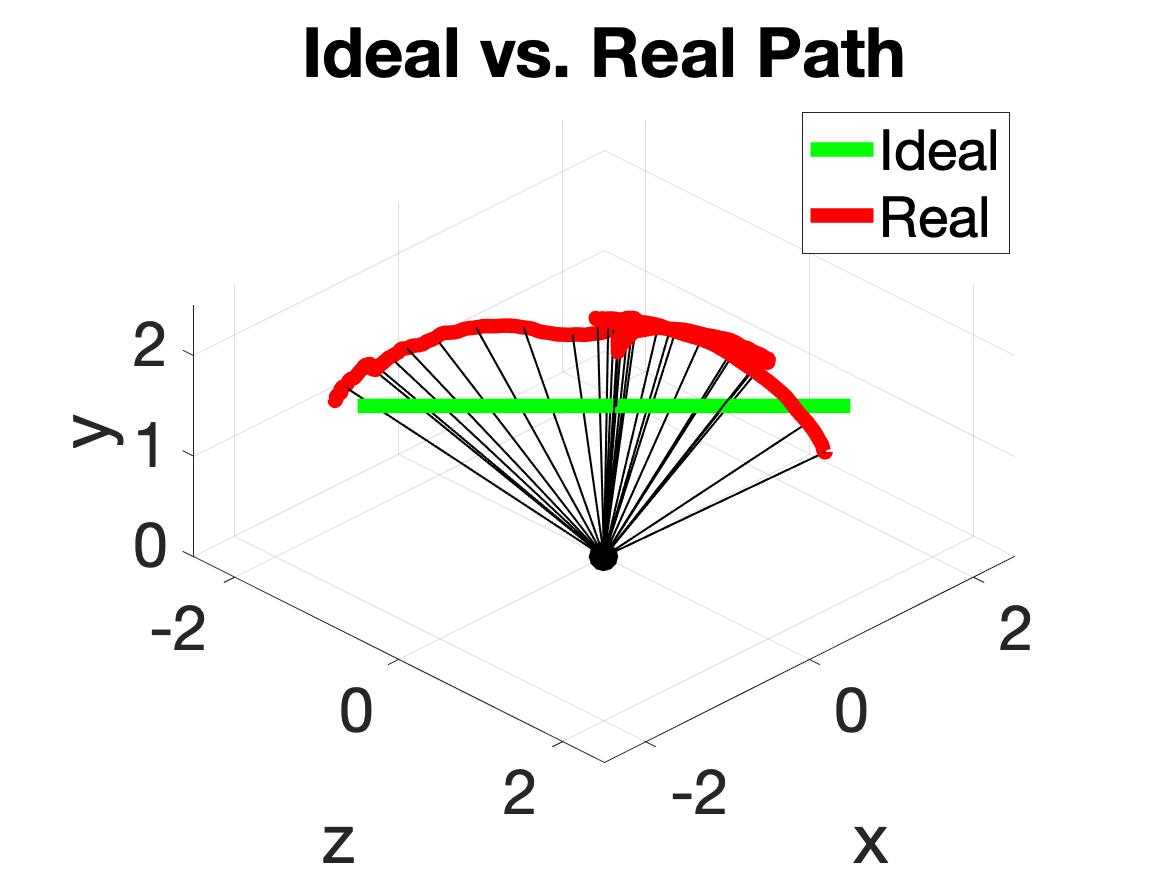}%
\label{fig::pos_15_2_2}}
\subfloat[Position 3m]{\includegraphics[width=0.4\columnwidth]{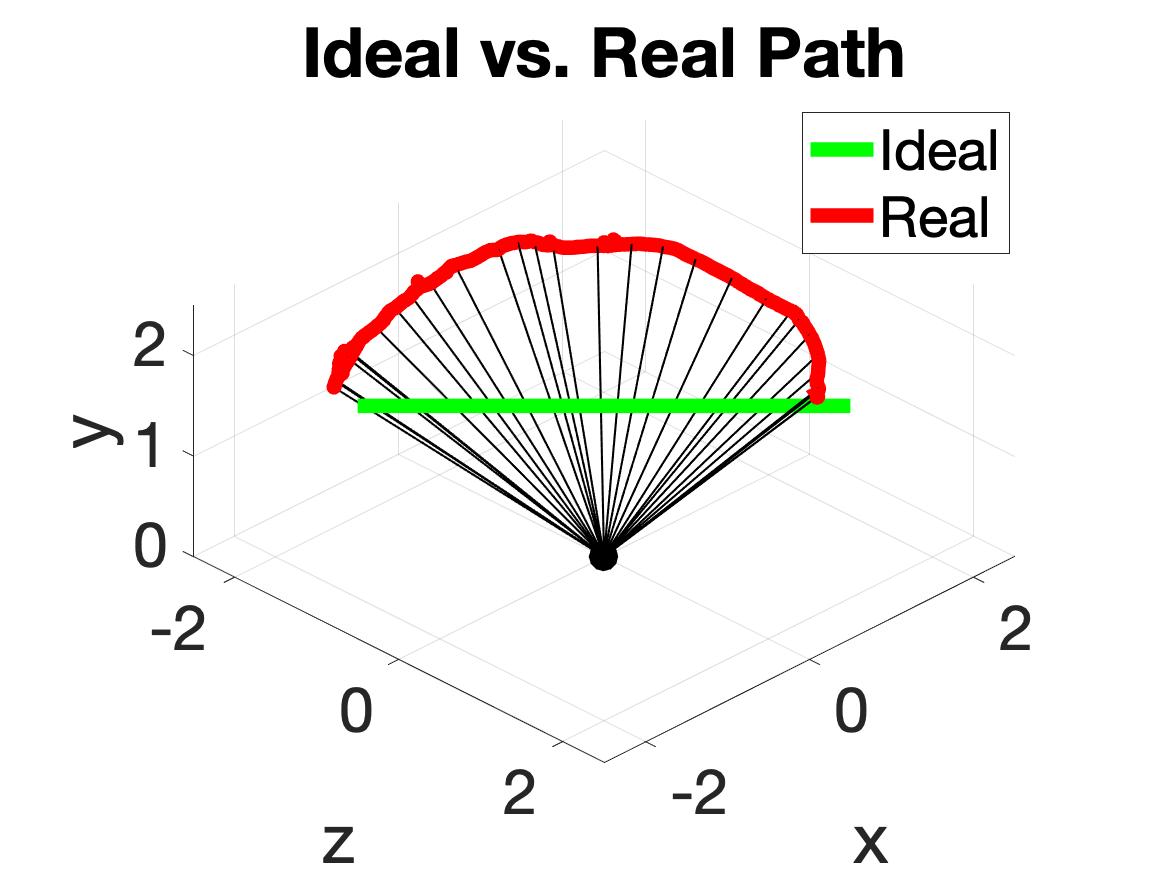}%
\label{fig::pos_3_1_2}}
\caption{Position Control Results}
\label{fig::experiments2}
\end{figure*}

Therefore, another set of experiment is conducted, with the focus on benchmarking the effect of waypoint density on position control accuracy. The path is designed to be horizontal and pass diagonally above the tether reel to amplify the effect of incoordination between control variables. Due to the singularity above the tether reel of the Jacobian matrix of velocity control, the UAV is inevitably trapped at the singularity when coming close to it. For velocity control, regions above the tether reel with 90\degree~elevation and indeterministic azimuth need to be avoided. Three position control trials are executed for each of the five waypoint densities, with one trial shown in Fig. \ref{fig::experiments2}.

\begin{figure}[]
\centering
\includegraphics[width=0.7\columnwidth]{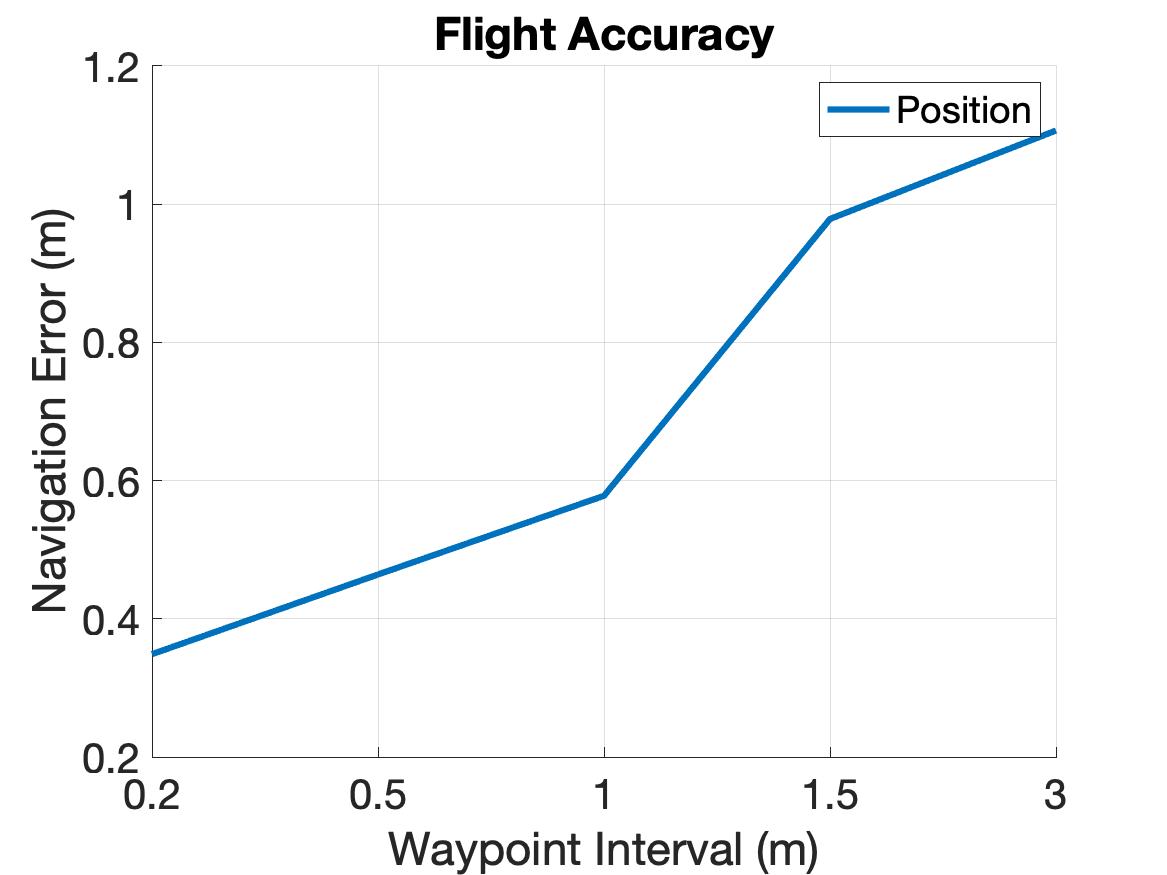}
\caption{Flight Accuracy for Path 2}
\label{fig::flight_accuracy2}
\end{figure}

In this set of experiments, it clearly shows that position control accuracy decreases with sparser waypoints (Fig. \ref{fig::flight_accuracy2}). From left to right in Fig. \ref{fig::experiments2}, the UAV deviates more and more from the ideal path, due to the lack of guidance between two consecutive waypoints. In the extreme case on the right hand side where only two waypoints denote the start and end position of the path, the UAV forms a semicircle-shaped trajectory instead of the intended straight line path. 

From the results of both experimental sets, position control works better with dense waypoints in terms of flight accuracy. This is because of the independent control over the three tether variables between waypoints. And denser waypoints provide extra guidance in between. However, denser waypoints also introduce jittery motion of the UAV since the shortsightedness causes overshoot so the path smoothness is no longer guaranteed. Proper waypoint density should be sufficiently dense to constrain the nondeterministic motion between waypoints while sparse enough to generate a smooth path. On the other hand, velocity control's accuracy is not very sensitive to the waypoint density, thanks to the coordination among the three tether variables using system's inverse Jacobian matrix. Similar to position control, smoothness of the path will be deteriorated by increasing waypoint density. Therefore, when using velocity control, sparse path plan is desirable as long as the critical points on the path is uniquely described by a minimum amount of waypoints. However, velocity controller could be trapped by singularity above the tether reel center, causing certain path to be not executable. Those areas need to be avoided when using velocity control only. An alternative approach is to use a composite controller which mostly uses velocity control but switches to position control when the UAV comes close to singularity. 

\section{CONCLUSIONS}
\label{sec::conclusions}
This paper presents two tether-based motion primitives to enable autonomous tethered UAV motion given pre-computed path plans. The two motion primitives are either based on three independent PID controllers or the system's inverse Jacobian matrix to compute control commands in the form of change rate of tether length, elevation, and azimuth angles. Both motion primitives are implemented on a tethered UAV in a MoCap studio, in order to benchmark their control performance with respect to different path plans. Path smoothness prefers paths with sparse waypoints for both motion primitives. However, position control's flight accuracy depends on proper waypoint density, which can provide extra guidance to minimize motion error between waypoints due to the independency of the three sub-controllers. The sparsity of waypoints is not an issue for the velocity control, thanks to the controller coordination enabled by the Jacobian matrix. But singularity exists for the velocity controller, where elevation angle is 90\degree~and azimuth is impossible to determine. Areas close to the top of the tether reel should be avoided using velocity control. A combination of both controllers to allow both accuracy and smoothness with sparse waypoints using velocity control and switchable to position control in the vicinity of singularity is suggested to enable high-quality tethered motion. Taking advantage of both motion primitives in order to improve tethered motion quality will be the focus of future work.

\section*{ACKNOWLEDGMENT}
This work is supported by NSF 1637955, NRI: A Collaborative Visual Assistant for Robot Operations in Unstructured or Confined Environments. 

\bibliographystyle{IEEEtran}
\bibliography{IEEEabrv,references}

\end{document}